\pdfoutput=1
\documentclass[11pt]{article}

\usepackage[final]{EMNLP2022}

\usepackage{times}
\usepackage{latexsym}
\usepackage{graphicx}
\usepackage{amsmath}
\usepackage{booktabs}
\usepackage{xcolor,colortbl}
\usepackage{hhline}
\usepackage{afterpage}
\usepackage[T1]{fontenc}
\usepackage[utf8]{inputenc}
\usepackage[greek, russian, english]{babel}

\usepackage{microtype}

\usepackage{inconsolata}

\usepackage{tabularx}

\usepackage{pifont}
\definecolor{darkpastelgreen}{rgb}{0.01, 0.75, 0.24}
\definecolor{darkpastelred}{rgb}{0.76, 0.23, 0.13}
\newcommand{\cmark}{\textcolor{darkpastelgreen}{\ding{51}}}%
\newcommand{\xmark}{\textcolor{darkpastelred}{\ding{55}}}%

\usepackage{rotating}

\title{\textsc{ACES}: Translation Accuracy Challenge Sets for Evaluating Machine Translation Metrics}

\author{Chantal Amrhein$^1$\thanks{{ } Equal contribution by all authors.} \and Nikita Moghe$^{2}$\footnotemark[1]  \and Liane Guillou$^{2}$\footnotemark[1] \\
  $^1$Department of Computational Linguistics, University of Zurich\\
  $^2$School of Informatics, University of Edinburgh \\ \medskip
  \texttt{amrhein@cl.uzh.ch, nikita.moghe@ed.ac.uk, lguillou@ed.ac.uk}}

\begin{document}
\maketitle
\begin{abstract}
As machine translation (MT) metrics improve their correlation with human judgement every year, it is crucial to understand the limitations of such metrics at the segment level. Specifically, it is important to investigate metric behaviour when facing accuracy errors in MT because these can have dangerous consequences in certain contexts (\textit{e.g.,} legal, medical). We curate \textsc{ACES}\footnote{Our dataset is available at \url{https://huggingface.co/datasets/nikitam/ACES} and the corresponding evaluation scripts at \url{https://github.com/EdinburghNLP/ACES}}, a Translation \textbf{A}ccuracy \textbf{C}halleng\textbf{E} \textbf{S}et, consisting of 68 phenomena ranging from simple perturbations at the word/character level to more complex errors based on discourse and real-world knowledge. We use \textsc{ACES} to evaluate a wide range of MT metrics including the submissions to the WMT 2022 metrics shared task and perform several analyses leading to general recommendations for metric developers. We recommend: a) combining metrics with different strengths, b) developing metrics that give more weight to the source and less to surface-level overlap with the reference and c) explicitly modelling additional language-specific information beyond what is available via multilingual embeddings.

\end{abstract}

\begin{figure*}
    \centering
    \includegraphics[width=\textwidth]{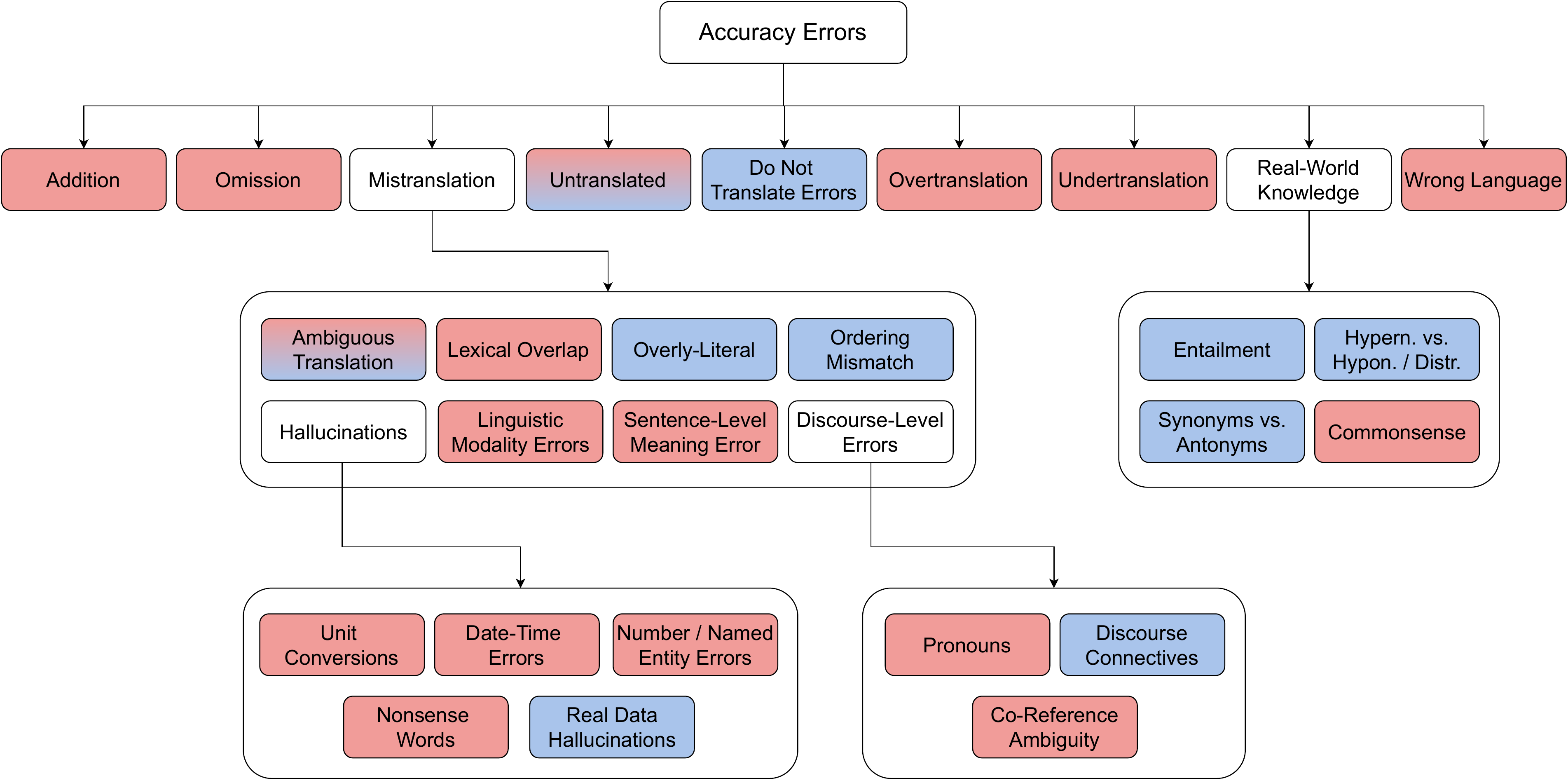}
    \caption{Diagram of the error categories on which our collection of challenge sets is based. Red means challenge sets are created automatically, blue means challenge sets are created manually.}
    \label{fig:diagram}
\end{figure*}

\section{Introduction}
Challenge sets have already been created for measuring the success of systems or metrics on a particular phenomenon of interest for a range of NLP tasks, including but not limited to: Sentiment Analysis\footnote{Submitted to the EMNLP 2017 ``Build It Break It'' shared task on sentiment analysis} \citep{li-etal-2017-bibi,mahler-etal-2017-breaking,staliunaite-bonfil-2017-breaking}, Natural Language Inference \citep{mccoy2019non,Rocchietti2021FANCYAD}, Question Answering \citep{ravichander-etal-2021-noiseqa}, Machine Reading Comprehension \citep{khashabi-etal-2018-looking}, Machine Translation (MT) \citep{king-falkedal-1990-using,isabelle-etal-2017-challenge}, and the more specific task of pronoun translation in MT \citep{guillou-hardmeier-2016-protest}. They are useful to compare the performance of different systems, or to identify performance improvement/degradation between a modified system and a previous iteration.

In this work, we describe the University of Zurich - University of Edinburgh submission to the \textit{Challenge Sets} subtask of the Conference on Machine Translation (WMT) 2022 Metrics shared task. Our Translation \textbf{A}ccuracy \textbf{C}halleng\textbf{E} \textbf{S}et (\textsc{ACES}) consists of 36,476 examples covering 146 language pairs and representing challenges from 68 phenomena (see Appendix~\ref{app:language_pair_matrix} for the distribution of examples across language pairs and Appendix~\ref{app:language_pair_phenomena} for the distribution of language pairs across phenomena). We focus on translation accuracy errors and base the phenomena covered in our challenge set on the Multidimensional Quality Metrics (MQM) ontology \citep{lommel2014}.
We include phenomena ranging from simple perturbations involving the omission/addition of characters or tokens, to more complex examples involving mistranslation e.g. ambiguity and hallucinations in translation, untranslated elements of a sentence, discourse-level phenomena, and real-world knowledge. 
We evaluate the metrics submitted to the WMT 2022 metrics shared task and a range of baseline metrics on \textsc{ACES}. Additionally, we perform an extensive analysis, which aims to reveal:

\begin{enumerate}
    \item The extent to which reference-based and reference-free metrics take into account the source sentence context.
    \item The extent to which reference-based metrics rely on surface-level overlap with the reference.
    \item Whether using multilingual embeddings results in better metrics.
\end{enumerate}

Based on our analysis, we recommend that metric developers consider: a) combining metrics with different strengths, e.g. in the form of ensemble models, b) paying more attention to the source and avoiding reliance on surface-overlap with the reference, and c) explicitly modelling additional language-specific information beyond what is available via multilingual embeddings. We also propose that \textsc{ACES} be used as a benchmark for developing evaluation metrics for MT to monitor which error categories can be identified better, and also whether there are any categories for which metric performance degrades. 

\section{Motivation}
With the advent of neural networks and especially Transformer-based architectures \citep{NIPS2017_7181}, machine translation outputs have become more and more fluent \citep{bentivogli-etal-2016-neural,toral-sanchez-cartagena-2017-multifaceted,article}. Fluency errors are also judged less severely than accuracy errors by human evaluators \citep{freitag-etal-2021-experts} which reflects the fact that accuracy errors can have dangerous consequences in certain contexts, for example in the medical and legal domains \citep{doi:10.1080/1369118X.2020.1776370}.

For these reasons, we decided to build a challenge set focused on accuracy errors. Specifically, we use the hierarchy of errors under the class \textit{Accuracy} from the MQM ontology to design these challenge sets. We extend this ontology by two error classes (translations defying real-world knowledge and translations in the wrong language) and specify several more specific subclasses such as discourse-level errors or ordering mismatches. A full overview of all error classes can be seen in Figure~\ref{fig:diagram}. Our challenge set consists of synthetically generated adversarial examples, examples from re-purposed contrastive MT test sets (both marked in red), and manually annotated examples (marked in blue). To create the challenge sets, we use test sets from tasks such as adversarial paraphrase detection, Natural Language Inference, and contrastive MT test sets created independently of the WMT shared tasks to avoid overlap with the data that is used to train neural evaluation metrics.

Another aspect we focus on is including a broad range of language pairs in \textsc{ACES}. Whenever possible we create examples for all language pairs covered in a source dataset when we use automatic approaches. For phenomena where we create examples manually, we also aim to cover at least two language pairs per phenomenon, but are of course limited to the languages spoken by the authors.

Finally, we aim to offer a collection of challenge sets covering both easy and hard phenomena. While it may be of interest to the community to continuously test on harder examples to check where machine translation evaluation metrics still break, we believe that easy challenge sets are just as important to ensure that metrics do not suddenly become worse at identifying error types that were previously considered ``solved''. Therefore, we take a holistic view when creating \textsc{ACES} and do not filter out individual examples or exclude challenge sets based on baseline metric performance or other factors.

We first discuss previous efforts to create challenge sets (Section~\ref{sec:related_work}), before giving a broad overview of the datasets used to construct \textsc{ACES} (Section~\ref{sec:datasets}) and discussing the individual challenge sets in more detail (Section~\ref{sec:challengesets}). We then introduce the metrics that participated in the shared task (Section~\ref{sec:eval_methodology}), present an overview of their performance on \textsc{ACES} (Section~\ref{sec: Results}) and detailed analyses (Section~\ref{sec:analysis}) that lead to a set of recommendations for future metric development (Section~\ref{sec:recommendations}).

\section{Related Work}
\label{sec:related_work}
Challenge sets are used to study a particular phenomenon of interest rather than the general distribution of phenomena in standard test sets \citep{popovic-castilho-2019-challenge}. The earliest introduction of challenge sets was by \citet{king-falkedal-1990-using} who probed acceptability of machine translations for different domains. Challenge sets have been prevalent in different fields within NLP such as parsing \citep{rimell-etal-2009-unbounded}, NLI \citep{mccoy2019non,Rocchietti2021FANCYAD}, question answering \citep{ravichander-etal-2021-noiseqa}, reading comprehension \citep{khashabi-etal-2018-looking} and sentiment analysis \citep{li-etal-2017-bibi,mahler-etal-2017-breaking,staliunaite-bonfil-2017-breaking}, to name a few. These challenge sets provide insights on whether state-of-the-art models are robust to domain shifts, and whether they have some understanding of linguistic phenomena like negation/commonsense or they simply rely on shallow heuristics. Another line of work under ``adversarial datasets'' also focuses on creating examples by perturbing the standard test set to fool the model (\citet{DBLP:journals/corr/abs-1207-0245, jia-liang-2017-adversarial},  \textit{inter-alia}).

Challenge sets for evaluating MT systems have focused on the translation models' ability to generate the correct translation given a phenomenon of interest. These include word sense ambiguity \citep{vamvas-sennrich-2021-contrastive}, gender bias \citep{rudinger-etal-2017-social, zhao-etal-2018-gender, stanovsky-etal-2019-evaluating}, structural divergence \citep{isabelle-etal-2017-challenge} and discourse level phenomena \citep{guillou-hardmeier-2016-protest,emelin-sennrich-2021-wino}.

While such challenge sets focus on evaluating specific machine translation models, it is necessary to identify whether the existing machine translation evaluation metrics also perform well under these and related phenomena. Developing challenge sets for machine translation metric evaluation has gained considerable interest because recently, neural MT evaluation metrics have shown improved correlation with human judgements \citep{freitag-etal-2021-results, kocmi-etal-2021-ship}. However, their weaknesses remain relatively unknown and only a small number of works (e.g. \citet {hanna-bojar-2021-fine} and \citet{amrhein2022identifying}) have proposed systematic analyses to uncover them. 

Previous challenge sets for metric evaluation focused on negation and sentiment polarity \citep{specia-etal-2020-findings} and synthetic perturbations such as antonym replacement, word omission, number swapping, punctuation removal, etc. \citep{freitag-etal-2021-results}. \citet{avramidis-etal-2018-fine} developed a manually constructed test suite of linguistically motivated perturbations for identifying weaknesses in reference-free evaluation. However, these challenge sets for metrics are only focused on high-resource language pairs such as English$\leftrightarrow$German and English$\rightarrow$Chinese. In this work, we repurpose existing machine translation challenge sets to evaluate machine translation evaluation metrics. We introduce several synthetically generated and manually created challenge sets that broadly focus on translation accuracy errors for 146 language pairs.

\section{Datasets}
\label{sec:datasets}
The majority of the examples in our challenge set were based on data extracted from three main datasets: FLORES-101, PAWS-X, and XNLI (with additional translations from XTREME).

The \textbf{FLORES-101} evaluation benchmark \citep{goyal-etal-2022-flores} consists of 3,001 sentences extracted from English Wikipedia and translated into 101 languages by professional translators. \textbf{FLORES-200} \citep{flores-200} expands the set of languages in FLORES-101. Originally intended for multilingual and low-resource MT evaluation, these datasets have a particular focus on low-resource languages.

\textbf{PAWS-X} \citep{yang-etal-2019-paws}, a cross-lingual dataset for paraphrase identification, consists of pairs of sentences that are labelled as true or adversarial paraphrases. It comprises the Wikipedia portion of the PAWS corpus \citep{zhang-etal-2019-paws} translated from English into six languages: French, Spanish, German, Chinese, Japanese, and Korean. The development and test sets (23,659 sentences total) were manually translated by professional translators, and the training set was translated using NMT systems via Google Cloud Translation\footnote{\url{https://cloud.google.com/translate}}.

\textbf{XNLI} \citep{conneau-etal-2018-xnli} is a multilingual Natural Language Inference (NLI) dataset consisting of 7,500 premise-hypothesis pairs with their corresponding inference label. The English examples were generated by crowd source workers before being manually translated into 14 languages: French, Spanish, German, Greek, Bulgarian, Russian, Turkish, Arabic, Vietnamese, Thai, Chinese, Hindi, Swahili and Urdu. In addition, we use the automatic translations from \textbf{XTREME} \citep{hu2020xtreme} of the XNLI test set examples from these 14 languages into English.

For the mistranslation phenomena
\hyperref[subsec:gender_in_occupation_names]{Gender in Occupation Names} and \hyperref[subsec:wsd]{Word Sense Disambiguation}, we leveraged the WinoMT and MuCoW datasets. \textbf{WinoMT} \citep{stanovsky-etal-2019-evaluating}, a challenge set developed for analysing gender bias in MT, contains 3,888 English examples extracted from the Winogender \citep{rudinger-etal-2017-social} and WinoBias \citep{zhao-etal-2018-gender} coreference test sets. WinoMT sentences cast participants into non-stereotypical gender roles and the dataset has an equal balance of male and female genders, and of stereotypical and non-stereotypical gender-role assignments (e.g., a female nurse vs. a female doctor). \textbf{MuCoW} \citep{raganato-etal-2019-mucow} is a multilingual contrastive, word sense disambiguation test suite for machine translation. The dataset covers 16 language pairs with more than 200,000 contrastive sentence pairs. It was automatically constructed from word-aligned parallel corpora and BabelNet's \citep{navigli-2012}  wide-coverage multilingual sense inventory. 

For the \hyperref[sec:discourse]{discourse-level phenomena}, we relied on \textit{annotated} resources developed specifically to support work on those phenomena in an MT setting. The \textbf{WMT 2018 English-German pronoun translation evaluation test suite} \citep{guillou-etal-2018-pronoun} contains 200 examples of the ambiguous English pronouns \textit{it} and \textit{they} extracted from the TED talks portion of ParCorFull \citep{lapshinova-koltunski-etal-2018-parcorfull}. The example sentences were translated into German by the 16 English-German systems submitted to WMT 2018, and the (German) pronoun translations were manually judged by human annotators as ``good/bad''. \textbf{Wino-X} \citep{emelin-sennrich-2021-wino} is a parallel dataset of German, French, and Russian Winograd schemas, aligned with their English counterparts. It was developed for commonsense reasoning and coreference resolution and used for this purpose to generate examples for  \hyperref[sec:commonsense-coref]{Commonsense Co-Reference Disambiguation}. The \textbf{Europarl ConcoDisco} corpus \citep{laali-kosseim-2017-improving} comprises the English-French parallel texts from Europarl \citep{koehn-2005-europarl} over which automatic methods were used to perform PDTB-style discourse connective annotation. Discourse connectives are labelled with their sense type and are aligned between the two languages.

\section{Challenge Sets}
\label{sec:challengesets}
Creating a contrastive challenge set for evaluating a machine translation evaluation metric requires a source sentence, a reference translation, and two translation hypotheses: one which contains an error or phenomenon of interest (the ``incorrect'' translation) and one which is a correct translation in that respect (the ``good'' translation). One possible way to create such challenge sets is to start with two alternative references (or two identical copies of the same reference) and insert errors into one of them to form an incorrect translation while the uncorrupted version can be used as the good translation. This limits the full evaluation scope to translation hypotheses that only contain a single error. To create a more realistic setup, we also create many challenge sets where the good translation is not free of errors, but it is a better translation than the incorrect translation. For automatically created challenge sets, we put measures in place to ensure that the incorrect translation is indeed a worse translation than the good translation.

\subsection{Addition and Omission}
\label{sec:addition-omission}
We create a challenge set for addition and omission errors which are defined in the MQM ontology as ``target content that includes content not present in the source'' and ``errors where content is missing from the translation that is present in the source'', respectively. We focus on the level of constituents and use an implementation by \citet{vamvas-sennrich-2022-little} to create synthetic examples of addition and omission errors.

To generate examples, we use the concatenated dev and devtest sets from the FLORES-101 evaluation benchmark. We focus on the 46 languages for which there exists a stanza parser\footnote{\url{https://stanfordnlp.github.io/stanza/available_models.html}} and create datasets for all languages paired with English plus ten additional language pairs that we selected randomly. The script by \citet{vamvas-sennrich-2022-little} randomly drops constituents from the source sentence and then generates two translations, one of the full source and one of the partial source without the constituent. Here is an example of two resulting translations:

\begin{small}
\vspace{0.5cm}
\setlength{\extrarowheight}{0.1cm}
\begin{tabularx}{0.95\columnwidth}{lX}
     Full: & For example, castle visits in the Loire Valley, the Rhine Valley, or a cruise \textbf{to interesting cities on the Danube} or \textbf{a} boat ride along the Erie Canal. \\
     Partial: & For example, castle visits in the Loire Valley, the Rhine Valley, or a cruise or boat ride along the Erie Canal. \vspace{0.35cm}
\end{tabularx}
\end{small}

Only partial translations that can be constructed by deleting spans from the full translation are considered. For translation, we use the M2M100\footnote{\url{https://huggingface.co/facebook/m2m100_1.2B}} model with 1.2B parameters \citep{fan2021beyond}.

We create \textbf{omission} examples by taking the original source and reference and using the translation of the full source as a good translation and the translation of the partial source as an incorrect translation. For \textbf{addition} errors, we test if the deleted span also occurs in the reference. If it doesn't, we discard the example, if it does, we delete that span from the reference and pair this partial reference with the partial source. Then, the good translation is the translation of the partial source and the incorrect translation is the translation of the full source. 
For language pairs with a BLEU score of less than 13 between the good translation and the reference, we manually check the examples to ensure the challenge set features appropriate examples of additions and omissions.

\subsection{Mistranslation - Ambiguous Translation}
\label{sec:source-disambig}
This error type is defined in the MQM ontology as a case where ``an unambiguous source text is translated ambiguously''. For this error type, we create challenge sets where MT metrics are presented with an unambiguous source and an ambiguous reference. The metrics then need to choose between two disambiguated translation hypotheses where only one meaning matches the source sentence. Therefore, these challenge sets test whether metrics consider the source when the reference is not expressive enough to identify the better translation. Since many reference-based metrics, by design, do not include the source to compute evaluation scores, we believe that this presents a challenging test set.

Our method for creating examples is inspired by \citet{vamvas-sennrich-2021-contrastive} who score a translation against two versions of the source sentence, one with an added correct disambiguation cue and one with a wrong disambiguation cue to determine whether a translation model produced the correct translation or not. Instead of adding the disambiguation cues to the source, we use an unambiguous source and add disambiguation cues to an ambiguous reference to create two contrasting translation hypotheses.

\subsubsection{Ambiguity - Occupation Names Gender}
\label{subsec:gender_in_occupation_names}
First, we create a challenge set based on WinoMT, where the challenge is to choose either a translation with a ``female'' or ``male'' disambiguation cue based on the source sentence:

\begin{small}
\vspace{0.5cm}
\setlength{\extrarowheight}{0.1cm}
\begin{tabular}{ll}
     SRC (de): &  Der Manager feuerte \textbf{die} Bäcker\textbf{in}. \\
     REF (en): & The manager fired the baker. \\
     \cmark: & The manager fired the \textbf{female} baker. \\
     \xmark: & The manager fired the \textbf{male} baker. \vspace{0.35cm}
\end{tabular}
\end{small}

We take all English sentences from the WinoMT dataset where either a pro-stereotypical or an anti-stereotypical occupation name occurs. The original sentences in WinoMT contain additional context from which the gender in the English sentence can be inferred. For example, the sentence above exists in the dataset once as ``The manager fired the baker because she was too rebellious.'' from which it is clear that the baker is female, and once as ``The manager fired the baker because he was upset.'' from which it is clear that the manager is male. To make the English sentences ambiguous, we remove the explanatory subordinate clauses using a sequence of regular expressions, so that the sentence becomes ``The manager fired the baker.'' where the gender of the manager and the baker are ambiguous.

We then add the disambiguation cues (``female'' or ``male'') to the ambiguous English sentences and translate them into German, French and Italian which are all languages that mark gender morphologically on most nouns that refer to a person. For translation, we use Google Translate\footnote{\url{https://translate.google.com/}} because we find that this system produces gendered occupation names that are largely faithful to the disambiguation cues. Finally, we remove explicit translations of ``female'' and ``male'' from the German, French or Italian output that would help the disambiguation beyond morphological cues. We predict the gender of the occupation names using the scripts provided by \citet{stanovsky-etal-2019-evaluating} and only keep translation pairs where both the translation of the male-disambiguated source is predicted to be male and the translation of the female-disambiguated source is predicted to be female. We then use either the German, French or Italian translation as the source sentence, the disambiguated English sentences as the translation candidates, and the ambiguous English sentence as the reference, as shown in the example above.

\subsubsection{Ambiguity - Word Sense Disambiguation}
\label{subsec:wsd}
Second, we create a challenge set based on MuCoW, where the challenge is to choose a translation with a sense-matching disambiguation cue based on the unambiguous source sentence:

\begin{small}
\vspace{0.5cm}
\setlength{\extrarowheight}{0.1cm}
\begin{tabular}{ll}
     SRC (de): &  Was heisst ``\textbf{Brühe}''?\\
     REF (en): & What does ``\textbf{stock}'' mean?\\
     \cmark: & What does ``\textbf{vegetable stock}'' mean?\\
     \xmark: & What does ``\textbf{penny stock}'' mean? \vspace{0.35cm}
\end{tabular}
\end{small}

We start with disambiguation cues that were automatically extracted by \citet{vamvas-sennrich-2021-contrastive} via masked language modelling. Initial screening of the data shows that some disambiguation cues are not sense-specific enough. Therefore, we decide to manually check all disambiguation cues and ensure they are sense-specific and if necessary, replace them with other cues. We generate three pairs of contrasting disambiguation cues per example and use the question ``What does X mean?'' as a pattern to create the challenge set examples. We decided against using sentences where ambiguous words occur naturally since it may be possible to infer the correct sense from the context of the English sentence rather than by looking at the unambiguous source word. We annotate each example as to whether the correct sense is the more frequent or less frequent sense using frequency counts provided by \citet{vamvas-sennrich-2021-contrastive}.
Following this methodology, we create challenge sets for German into English and Russian into English.

\subsubsection{Ambiguity - Discourse Connectives}
\label{subsec:discourse_connectives}
Third, we create a challenge set where the challenge is to identify a translation with the correct discourse connective based on the unambiguous source sentence:

\begin{small}
\vspace{0.5cm}
\setlength{\extrarowheight}{0.1cm}
\begin{tabularx}{0.95\columnwidth}{lX}
     SRC (fr): & Aucun test de qualité de l'air n'ait été réalisé dans ce bâtiment \textbf{depuis} notre élection. \\
     REF (en): & No air quality test has been done on this particular building \textbf{since} we were elected. \\
     \cmark: & No air quality test has been done on this particular building \textbf{from the time} we were elected. \\
     \xmark: & No air quality test has been done on this particular building \textbf{because} we were elected. \vspace{0.35cm}
\end{tabularx}
\end{small}

The English discourse connective ``since'' can have either causal or temporal meaning, which is expressed explicitly in both French and German. Exploiting this fact, we use the ambiguous ``since'' in the reference and create two contrastive translations one with ``because'' for causal meaning and one with ``from the time'' for temporal meaning. The correct translation is determined by looking at the French or German source sentence where this information is marked explicitly. We use the discourse connective annotations in the Europarl ConcoDisco corpus for this challenge set. We use an automatic-guided search based on the French discourse connective ``depuis'' (which has temporal meaning) to identify candidate translation pairs. We then manually construct valid contrasting examples for causal and temporal ``since'' based on the English reference. This results in a challenge set for French-English but we also create a German-English version of the challenge set, where we translate the French source sentences into German and manually correct them.

\subsection{Mistranslation - Hallucinations}
\label{sec:hallucination}
In this category, we group together several subcategories of mistranslation errors that happen at the word level and could occur due to hallucination by an MT model. Such errors are \hyperref[subsec:units]{wrong units}, \hyperref[subsec:date-time]{wrong dates or times}, \hyperref[subsec:levels]{wrong numbers or named entities}, as well as \hyperref[sec:nonsense]{hallucinations at the subword level} that result in nonsensical words. We also present a challenge set of annotated \hyperref[subsec:real_hallucination]{hallucinations in real MT outputs}. These challenge sets test whether the machine translation evaluation metrics can reliably identify hallucinations when presented with a correct alternative translation.

\subsubsection{Hallucination - Date-Time Errors}
\label{subsec:date-time}
We create a challenge set for the category of ``date-time errors''. To do this, we collect month names and their abbreviations for several language pairs. We then form a good translation by swapping a month's name with its abbreviation. The corresponding incorrect translation is generated by swapping the month name with another month name:

\begin{small}
\vspace{0.5cm}
\setlength{\extrarowheight}{0.1cm}
\begin{tabularx}{0.95\columnwidth}{lX}
     SRC (pt): & Os manifestantes esperam coletar uma petição de 1,2 milhão de assinaturas para apresentar ao Congresso Nacional em \textbf{novembro}. \\
     REF (en): & Protesters hope to collect a petition of 1.2 million signatures to present to the National Congress in \textbf{November}. \\
     \cmark: & The protesters expect to collect a petition of 1.2 million signatures to be submitted to the National Congress in \textbf{Nov.} \\
     \xmark: & The protesters expect to collect a petition of 1.2 million signatures to be submitted to the National Congress in \textbf{August}. \vspace{0.35cm}
\end{tabularx}
\end{small}

To create this dataset, we use the automatic translations of the FLORES-101 dataset from Section~\ref{sec:addition-omission}. We choose all pairs with target languages for which we know the abbreviations for months\footnote{\url{https://web.library.yale.edu/cataloging/months}} which results in 70 language pairs. As a measure of control, we check that the identified month names in the translation also occur in the reference. If they do not, we exclude the example.

\subsubsection{Hallucination - Numbers and Named Entities}
\label{subsec:levels}
We create a challenge set for numbers and named entities where the challenge is to identify translations with incorrect numbers or named entities. Following the analysis by \citet{amrhein2022identifying}, we perform character-level edits (adding, removing or substituting digits in numbers or characters in named entities) as well as word-level edits (substituting whole numbers or named entities). In the 2021 WMT metrics shared task, number differences were not a big issue for most neural metrics \citep{freitag-etal-2021-results}. However, we believe that simply changing a number in an alternative translation and using this as an incorrect translation as done by \citet{freitag-etal-2021-results} is an overly simplistic setup and does not cover the whole translation hypothesis space.

To address this shortcoming, we propose a three-level evaluation (see examples below). The first, easiest level follows \citet{freitag-etal-2021-results} and applies a change to an alternative translation to form an incorrect translation. The second level uses an alternative translation that is lexically very similar to the reference as the good translation and applies a change to the reference to form an incorrect translation. The third, and hardest level, uses an alternative translation that is lexically very different from the reference as the good translation and applies a change to the reference to form an incorrect translation. In this way, our challenge set tests whether number and named entity differences can still be detected as the surface similarity between the two translation candidates decreases and the surface similarity between the incorrect translation and the reference increases.

\begin{small}
\vspace{0.5cm}
\begin{tabularx}{0.95\columnwidth}{lX}
     SRC (es): & Sin embargo, Michael Jackson, Prince y \textbf{Madonna} fueron influencias para el álbum. \\
     REF (en): & Michael Jackson, Prince and \textbf{Madonna} were, however, influences on the album. \\\\\hline\\
    Level-1 \cmark: & However, Michael Jackson, Prince, and \textbf{Madonna} were influences on the album. \\
    Level-1 \xmark: & However, Michael Jackson, Prince, and \textbf{Garza} were influences on the album. \\\\\hline\\
    Level-2 \cmark: & However, Michael Jackson, Prince, and \textbf{Madonna} were influences on the album. \\
    Level-2 \xmark: &  Michael Jackson, Prince and \textbf{Garza} were, however, influences on the album.\\\\\hline\\
    Level-3 \cmark: & The record was influenced by \textbf{Madonna}, Prince, and Michael Jackson  though. \\
    Level-3 \xmark: & Michael Jackson, Prince and \textbf{Garza} were, however, influences on the album.\vspace{0.35cm}
\end{tabularx}
\end{small}

We use cross-lingual paraphrases from the PAWS-X dataset as a pool of alternative translations to create this challenge set. For levels 2 and 3, we measure surface-level similarity with Levenshtein distance\footnote{\url{https://github.com/life4/textdistance}} at the character-level and use spacy\footnote{\url{https://spacy.io/}} \citep{spacy} for identifying named entities of type ``person''. To substitute whole named entities, we make use of the \texttt{names}\footnote{\url{https://github.com/treyhunner/names}} Python library. We only consider language pairs for which we can use a spacy NER model on the target side, which results in 42 language pairs.

\subsubsection{Hallucination - Unit Conversion}
\label{subsec:units}
We create a challenge set for unit conversions where the challenge is to identify the correct unit conversion:

\begin{small}
\vspace{0.5cm}
\setlength{\extrarowheight}{0.1cm}
\begin{tabularx}{0.95\columnwidth}{lX}
     SRC (de): & Auf einem \textbf{100 Fuß} langen Teilabschnitt läuft Wasser über den Damm. \\
     REF (en): & Water is spilling over the levee in a section \textbf{100 feet} wide. \\
     \cmark: & On a \textbf{30.5 metres} long section, water flows over the dam. \\
     \xmark: & On a \textbf{100 metres} long section, water flows over the dam. \vspace{0.35cm}
\end{tabularx}
\end{small}

We take all source sentences, reference sentences and translations of the FLORES-101 sets from Section~\ref{sec:addition-omission}. We only use the 45 language pairs into English since the Python packages we use for unit conversion only work for English. We first use the Python package \texttt{quantulum3}\footnote{\url{https://github.com/nielstron/quantulum3}} to extract unit mentions from text. We only consider sentences where we identify the same unit mentions in the translation as in the reference and we remove self-disambiguating unit mentions, like ``645 miles (1040 km)'' from the reference and translation. Then, we use the Python package \texttt{pint}\footnote{\url{https://github.com/hgrecco/pint}} to convert unit mentions in the translation into different units. The permitted conversions are listed in Appendix~\ref{app:allowed-units}. 

The sentence with the converted amount and new unit is considered to be the good translation. Based on this sentence, we construct two incorrect versions, one where the amount matches the reference but the unit is still converted (see example above) and one where the amount is the converted amount but the unit is copied from the reference. We pair each incorrect translation with the good translation and add both examples to the challenge set individually. We are aware that this challenge set lies beyond the ability of current MT systems and evaluation metrics, however, we believe challenge sets such as these incentivise future work on such capabilities which would reduce the workload in post-editing.

\subsubsection{Hallucination - Nonsense Words}
\label{sec:nonsense}
We also consider more natural hallucinations at the subword level. Because recent MT systems are trained with subwords \citep{sennrich-etal-2016-neural}, an MT model may choose a wrong subword at a specific time step such that the resulting token is not a known word in the target language. With this challenge set, we are interested in how well neural MT evaluation metrics that incorporate subword-level tokenisation can identify such ``nonsense'' words.

To create this challenge set, we consider tokens which are broken down into at least two subwords and then randomly swap those subwords with other subwords to create nonsense words. In the example below, ``mass'' is broken down as ``mas'' and ``\#\#s'' using subwords and the new word is created by swapping ``mas'' with ``in'' while retaining ``\#\#s'', creating ``ins'' as the nonsense word. We use the paraphrases from the PAWS-X dataset as good translations and randomly swap one subword in the reference to generate an incorrect translation. This perturbation is language-agnostic. We use the multilingual BERT \citep{devlin-etal-2019-bert} tokeniser to replace the subwords.

\begin{small}
\vspace{0.5cm}
\setlength{\extrarowheight}{0.1cm}
\begin{tabularx}{0.95\columnwidth}{lX}
     SRC (de): & Die \textbf{Massen}produktion von elektronischen und digitalen Filmen war bis zum Aufkommen der pornographischen Videotechnik direkt mit der Mainstream-Filmindustrie verbunden. \\
     REF (en): &	The \textbf{mas}s production of electronic and digital films was directly linked to the mainstream film industry until the emergence of pornographic video technology.\\
     \cmark: & Until the advent of pornographic video technology , the mass production of electronic and digital films was tied directly to the mainstream film industry.\\
     \xmark: & The \textbf{in}s production of electronic and digital films was directly linked to the mainstream film industry until the emergence of pornographic video technology. \vspace{0.35cm}
\end{tabularx}
\end{small}

\subsubsection{Hallucination - Real Data Hallucinations}
\label{subsec:real_hallucination}
The previously discussed hallucination challenge sets were all created automatically. In addition to these challenge sets, we also create one with real data hallucinations.

For this dataset, we manually check the translations of the FLORES-101 dev and devtest sets for four language pairs: de$\rightarrow$en, en$\rightarrow$de, fr$\rightarrow$de and en$\rightarrow$mr. We consider both cases where a more frequent, completely wrong word occurs and cases where the MT model started with the correct subword but then produced random subwords as hallucinations. Translations with a hallucination are used as incorrect translations. We manually replace the hallucination part with its correct translation to form the good translation. If possible, we create one good translation by copying the corresponding token(s) from the reference and one with a synonymous token that does not match the reference:

\begin{small}
\vspace{0.5cm}
\setlength{\extrarowheight}{0.1cm}
\begin{tabularx}{0.95\columnwidth}{lX}
     SRC (de): & Es wird angenommen, dass dieser voll gefiederte warmblütige Raubvogel aufrecht auf zwei Beinen lief und \textbf{Krallen} wie der Velociraptor hatte. \\
     REF (en): & This fully feathered, warm blooded bird of prey was believed to have walked upright on two legs with \textbf{claws} like the Velociraptor.\\
     \cmark{} (copy): & It is believed that this fully feathered warm-blooded predator ran upright on two legs and had \textbf{claws} like the Velociraptor.\\
     \cmark{} (syn.): & It is believed that this fully feathered warm-blooded predator ran upright on two legs and had \textbf{talons} like the Velociraptor.\\
     \xmark: & It is believed that this fully feathered warm-blooded predator ran upright on two legs and had \textbf{crumbs} like the Velociraptor. \vspace{0.35cm}
\end{tabularx}
\end{small}

\subsection{Mistranslation - Lexical Overlap}
\label{subsec:lexical-overlap}
Language models trained with the masked language modelling objective are successful on downstream tasks because they model higher-order word co-occurrence statistics instead of syntactic structures \citep{sinha-etal-2021-masked}. Although this has been shown for a monolingual English model, we expect that multilingual pre-trained models, as well as MT metrics finetuned on such models, exhibit such behaviour. Similarly, existing surface-level metrics rely on n-gram matching between the hypothesis and the reference. Thus, we are interested in whether MT evaluation metrics can reliably identify the incorrect translation if it shares a high degree of lexical overlap with the reference:

\begin{small}
\vspace{0.5cm}
\setlength{\extrarowheight}{0.1cm}
\begin{tabularx}{0.95\columnwidth}{lX}
     SRC (fr): & En 1924, il a été porte-parole invité de l'ICM à Toronto, à Oslo en 1932 et à Zurich en 1936. \\
     REF (en): & In 1924 he was an invited spokesman for the ICM in Toronto, in \textbf{Oslo in 1932} and in \textbf{1936 in Zurich.}\\
     \cmark: & He served as a guest speaker for ICM in 1924, 1932 and 1936 in Toronto, Oslo and Zurich.\\
     \xmark: & He was an invited spokesman for the ICM in Toronto in 1924, in \textbf{Zurich in 1932} and in \textbf{Oslo in 1936.} \vspace{0.35cm}
\end{tabularx}
\end{small}

In this example, Oslo and Zurich are swapped in the ``incorrect translation'' making the sentence factually incorrect. To create such examples, we use the PAWS-X dataset for which adversarial paraphrase examples were constructed by changing the word order and/or the syntactic structure while maintaining a high degree of lexical overlap. We only consider examples in the development set that are adversarial paraphrases.

We automatically translate the first example in a pair (fr$\rightarrow$en, en$\rightarrow$fr, en$\rightarrow$ja) and then manually correct the translations for en, fr, and ja to obtain 100 ``good translations'' per language. We use the corresponding first paraphrase as the ``reference'' and the second (adversarial) paraphrase as the ``incorrect translation''. We then pair these examples with the first paraphrase in the remaining six languages in PAWS-X to obtain the ``source''. Following this methodology we create examples for each target language (xx$\rightarrow$en, xx$\rightarrow$fr, xx$\rightarrow$ja).

\subsection{Mistranslation - Linguistic Modality}
Modal auxiliary verbs signal the function of the main verb that they govern. For example, they may be used to denote possibility (``could''), permission (``may''), the giving of advice (``should''), or necessity (``must''). We are interested in whether MT evaluation metrics can identify when modal auxiliary verbs are incorrectly translated:

\begin{small}
\vspace{0.5cm}
\setlength{\extrarowheight}{0.1cm}
\begin{tabularx}{0.95\columnwidth}{lX}
     SRC (de): & Mit der Einführung dieser Regelung \textbf{könnte} diese Freiheit enden.\\
     REF (en): & With this arrangement in place, this freedom \textbf{might} end.\\
     \cmark{}: & With the introduction of this regulation, this freedom \textbf{could} end.\\
     \xmark{}: & With the introduction of this regulation, this freedom \textbf{will} end.
     \vspace{0.35cm}
\end{tabularx}
\end{small}

We focus on the English modal auxiliary verbs: ``must'' (necessity), and ``may'', ``might'', ``could'' (possibility). We begin by identifying parallel sentences where there is a modal verb in the German source sentence and one from our list (above) in the English reference. We then translate the source sentence using Google Translate to obtain the ``good'' translation and manually replace the modal verb with an alternative with the same meaning where necessary (e.g. ``have to'' denotes necessity as does ``must''; also ``might'', ``may'' and ``could'' are considered equivalent). For the incorrect translation, we manually substitute the modal verb that conveys a different meaning or \textit{epistemic strength} e.g. in the example above ``might'' (possibility) is replaced with ``will'', which denotes (near) certainty. Instances of ``may'' with \textit{deontic} meaning (e.g. expressing permission) are excluded from the set, leaving only those with an \textit{epistemic} meaning (expressing probability or prediction). We also construct examples in which the modal verb is omitted from the incorrect translation.

We employ two strategies to create examples: one in which the modal auxiliary is substituted, and another where it is deleted. We use a combination of the FLORES-200 and PAWS-X datasets as the basis of the challenge sets.

\subsection{Mistranslation - Overly Literal Translations}

MQM defines this error type as translations that are overly literal, for example literal translations of figurative language. Here, we look specifically at \hyperref[subsec:idioms]{idioms} and at \hyperref[subsec:real_overly_literal]{real-data errors}.

\subsubsection{Overly Literal - Idioms}
\label{subsec:idioms}
Idioms tend to be translated overly literally \citep{dankers-etal-2022-transformer} and it is interesting to see if such translations are also preferred by neural machine translation evaluation metrics, which likely have not seen many idioms during finetuning:

\begin{small}
\vspace{0.5cm}
\setlength{\extrarowheight}{0.1cm}
\begin{tabularx}{0.95\columnwidth}{lX}
     SRC (de): & Er hat versucht, mir die Spielregeln zu erklären, aber \textbf{ich verstand nur Bahnhof}. \\
     REF (en): & He tried to explain the rules of the game to me, but \textbf{I did not understand them}. \\
     \cmark: & He tried to explain the rules of the game to me, but \textbf{it was all Greek to me}. \\
     \xmark: & He tried to explain the rules of the game to me, but \textbf{I only understood train station}. \vspace{0.35cm}
\end{tabularx}
\end{small}

We create this challenge set based on the PIE\footnote{\url{https://github.com/zhjjn/MWE_PIE}} parallel corpus of English idiomatic expressions and literal paraphrases \citep{zhou-etal-2021-pie}. We manually translate 102 parallel sentences into German for which we find a matching idiom that is not a word-by-word translation of the original English idiom. Further, we create an overly-literal translation of the English and German idioms. We use either the German or English original idiom as the source sentence. Then, we either use the correct idiom in the other language as the reference and the literal paraphrase as the good translation, or vice versa. The incorrect translation is always the overly-literal translation of the source idiom.

\subsubsection{Overly-Literal - Real Data Errors}
\label{subsec:real_overly_literal}

We are also interested in overly-literal translations occurring in real data: 

\begin{small}
\vspace{0.5cm}
\setlength{\extrarowheight}{0.1cm}
\begin{tabularx}{0.95\columnwidth}{lX}
     SRC (de): & Today, the only insects that cannot fold back their wings are \textbf{dragon flies} and mayflies. \\
     REF (en): & Heute sind \textbf{Libellen} und Eintagsfliegen die einzigen Insekten, die ihre Flügel nicht zurückklappen können. \\
     \cmark{} (copy) : & Heute sind die einzigen Insekten, die ihre Flügel nicht zurückbrechen können, \textbf{Libellen} und Mayflies. \\
     \cmark{} (syn.): & Heute sind die einzigen Insekten, die ihre Flügel nicht zurückbrechen können, \textbf{Wasserjungfern} und Mayflies. \\
     \xmark: & Heute sind die einzigen Insekten, die ihre Flügel nicht zurückbrechen können, \textbf{Drachenfliegen} und Mayflies. \vspace{0.35cm}
\end{tabularx}
\end{small}

For this challenge set, we manually check MT translations of the FLORES-101 datasets. If we find an overly-literal translation, we manually correct it to form the good translation. We create one good translation where we copy the part of the reference that corresponds to the overly-literal part and, if possible, another good translation where we use a synonym of the reference token. This challenge set contains examples for four language pairs: de$\rightarrow$en, en$\rightarrow$de, fr$\rightarrow$de and en$\rightarrow$mr.

\subsubsection{Mistranslation - Sentence-Level Meaning Error}
\label{subsec:lexically-similar}
We also consider a special case of sentence-level semantic error that arises due to the nature of the task of Natural Language Inference (NLI). The task of NLI requires identifying where the given hypothesis is an entailment, contradiction, or neutral, with respect to a given premise. As a result, the premise and hypothesis have substantial overlap but they vary in meaning. We are interested in whether MT evaluation metrics can pick up on such sentence-level meaning changes:

\begin{small}
\vspace{0.5cm}
\setlength{\extrarowheight}{0.1cm}
\begin{tabularx}{0.95\columnwidth}{lX}
     SRC (el): &  \foreignlanguage{greek}{Ο πραγματικός θόρυβος ελκύει τους ηλικιωμένους}.\\
     REF (en): & Real noise appeals to the old. (premise) \\
     \cmark: & The real noise attracts the elderly. \\
     \xmark: &  Real noise appeals to the young and appalls the old. (hypothesis)\vspace{0.35cm}
\end{tabularx}
\end{small}

We use the XNLI dataset to create such examples. We consider examples where there is at least 0.5 chrF score between the English premise and hypothesis and where the labels are either contradiction or neutral. Examples with an entailment label are excluded as some examples in the dataset are paraphrases of each other and there would be no sentence-level meaning change. We discuss effects of entailment in Section~\ref{sec:real_world_knowledge_entailment}. We use either the premise or the hypothesis as the reference and an automatic translation as the ``good translation''. The corresponding premise or hypothesis from the remaining 14 languages is used as the source. The ``incorrect translation'' is either the premise if the reference is the hypothesis, or vice versa.

\subsection{Mistranslation - Ordering Mismatch}
We also investigate the effects of changing word order in a way that changes meaning:

\begin{small}
\vspace{0.5cm}
\setlength{\extrarowheight}{0.1cm}
\begin{tabularx}{0.95\columnwidth}{lX}
     SRC (de): & Erfülle Dein Zuhause mit einem köstlichem \textbf{Kaffee} am Morgen und etwas entspannendem \textbf{Kamillentee} am Abend. \\
     REF (en): & Fill your home with a rich \textbf{coffee} in the morning and some relaxing \textbf{chamomile tea} at night. \\
     \cmark: & Fill your home with a delicious \textbf{coffee} in the morning and some relaxing \textbf{chamomile tea} in the evening. \\
     \xmark: & Fill your home with a delicious \textbf{chamomile tea} in the morning and some relaxing \textbf{coffee} in the evening. \vspace{0.35cm}
\end{tabularx}
\end{small}

This challenge set is created manually by changing translations from the FLORES-101 dataset and covers de$\rightarrow$en, en$\rightarrow$de and fr$\rightarrow$de.

\subsection{Mistranslation - Discourse-level Errors}
\label{sec:discourse}
We introduce a new subclass of mistranslation errors that specifically cover discourse-level phenomena.
\subsubsection{Discourse-level Errors - Pronouns}

First, we are interested in how MT evaluation metrics handle various discourse-level phenomena related to pronouns. To create these challenge sets, we use the English-German pronoun translation evaluation test suite from the WMT 2018 shared task as the basis for our examples.

We extract all translations (by the English-German WMT 2018 systems) that were marked as ``correct'' by the human annotators, for the following six categories derived from the manually annotated pronoun function and attribute labels: pleonastic \textit{it}, anaphoric subject and non-subject position \textit{it}, anaphoric \textit{they}, singular \textit{they}, and group \textit{it/they}. In the case of anaphoric pronouns, we select only the inter-sentential examples (i.e. where the sentence contains both the pronoun and its antecedent). We use the MT translations as the ``good'' translations and automatically generate ``incorrect'' translations using one of the following strategies: \textit{omission} - the translated pronoun is deleted from the MT output, \textit{substitution} - the ``correct'' pronoun is replaced with an ``incorrect'' form.

For \textit{anaphoric} pronouns, when translated from English into a language with grammatical gender, such as German, the pronoun translation must a) agree in number and gender with the translation of its antecedent, and b) have the correct grammatical case. We propose ``incorrect'' translations as those for which this agreement does not hold:

\begin{small}
\vspace{0.5cm}
\setlength{\extrarowheight}{0.1cm}
\begin{tabular}{ll}
     SRC (en): & I have a \textit{shopping bag}; \textbf{\textit{it}} is red.\\
     REF (de): & Ich habe eine \textit{Einkaufstüte}; \textbf{\textit{sie}} ist rot.\\
     \cmark: & Ich habe einen \textit{Einkaufsbeutel}; \textbf{\textit{er}} ist rot.\\
     \xmark{} (subs.): & Ich habe einen \textit{Einkaufsbeutel}; \textbf{\textit{sie}} ist rot.\\
     \xmark{} (omit): & Ich habe einen \textit{Einkaufsbeutel}; \textbf{\textit{\O}} ist rot.
     \vspace{0.35cm}
\end{tabular}
\end{small}

Conversely, for \textit{pleonastic} uses of ``it'' no agreement is required, instead, the correct translation in German requires a simple mapping: ``it'' $\rightarrow$ ``es''. An `incorrect'' translation of pleonastic `it' in German could be ``er'' (masc. sg.) or ``sie'' (fem. sg., or pl.). We create, for each ``correct'' translation a set of possible ``incorrect'' values and automatically select one  at random to replace the ``correct'' pronoun. For example, in the pleonastic case:

\begin{small}
\vspace{0.5cm}
\setlength{\extrarowheight}{0.1cm}
\begin{tabular}{ll}
     SRC (en): & \textbf{It} is raining\\
     REF (de): & \textbf{Es} regnet\\
     \cmark{}: & \textbf{Es} regnet\\
     \xmark{} (subs.): & \textbf{Er} regnet\\
     \xmark{} (omit): & \textbf{\O{}} regnet
\end{tabular}
\vspace{0.35cm}
\end{small}

\subsubsection{Discourse-level Errors - Discourse Connectives}
The English discourse connective ``while'' is ambiguous -- it may be used with either a \textit{Comparison.Contrast} or \textit{Temporal.Synchrony} sense -- as are two of its possible translations into French: ``tandis que'' and ``alors que''. We leverage a corpus of parallel English/French sentences with discourse connectives marked and annotated for sense, and select examples with ambiguity in the French source sentence. We construct the good translation by replacing instances of ``while'' temporal with ``as'' or ``as long as'' and instances of ``while'' comparison as ``whereas'' (ensuring grammaticality is preserved). For the incorrect translation, we replace the discourse connective with one with the alternative sense of ``while'' e.g. we use ``whereas'' (comparison) where a temporal sense is required:

\begin{small}
\vspace{0.5cm}
\setlength{\extrarowheight}{0.1cm}
\begin{tabularx}{0.95\columnwidth}{lX}
     SRC (fr): & Dans l'UE-10, elles ont progressé de 8\% \textbf{tandis que} la dette pour l'UE-2 a augmenté de 152\%. \\
     REF (en): & In EU-10 they grew by 8\% \textbf{while} the debt for the EU-2 increased by 152\%. \\
     \cmark: & In the EU-10, they increased by 8\% \textbf{when} the debt for the EU-2 increased by 152\%. \\
     \xmark: & In the EU-10, they increased by 8\% \textbf{whereas} the debt for the EU-2 increased by 152\%. \vspace{0.35cm}
\end{tabularx}
\end{small}

We extract our examples from the Europarl ConcoDisco dataset. We automatically selected the sentence pairs that contain an instance of ``while'' in English and either ``alors que'' or ``tandis que'' in French. Our dataset contains examples for both the \textit{Comparison.Contrast} sense and the \textit{Temporal.Synchrony} sense.

This challenge set complements the discourse connectives set in section~\ref{subsec:discourse_connectives}, in which the English discourse connective ``since'' is ambiguous, but the corresponding connectives in French and German are not. Note that while in the previous challenge set the correct translation can be identified by looking at the source, here metrics can only rely on context to identify the correct discourse connective.

\subsubsection{Discourse-level Errors - Commonsense Co-Reference Disambiguation}
\label{sec:commonsense-coref}
One of the greater challenges within computational coreference resolution is referring to the correct antecedent by using commonsense/real-world knowledge. \citet{emelin-sennrich-2021-wino} construct a benchmark to test whether multilingual language models and neural machine translation models can perform such commonsense coreference resolutions. We are interested in whether such commonsense coreference resolutions pose a challenge for MT evaluation metrics:

\begin{small}
     \vspace{0.5cm}
\begin{tabularx}{0.95\columnwidth}{lX}
\setlength{\extrarowheight}{0.1cm}

     SRC (en): & It took longer to clean the fish tank than the dog cage because \textbf{it} was dirtier.	\\
     REF (de): & Das Reinigen des Aquariums dauerte länger als das des Hundekäfigs, da \textbf{es} schmutziger war.\\
     \cmark : & Das Reinigen des Aquariums dauerte länger als das des Hundekäfigs, da \textbf{das Aquarium} schmutziger war.		\\
    
     \xmark{} : & Die Reinigung des Aquariums dauerte länger als die des Hundekäfigs, da \textbf{er} schmutziger war.	
     \vspace{0.35cm}
\end{tabularx}
\end{small}

The English sentences in the Wino-X challenge set were sampled from the Winograd schema. All contain the pronoun \textit{it} and were manually translated into two contrastive translations for de, fr, and ru. Based on this data, we create our challenge sets covering two types of examples: For the first, the good translation contains the pronoun referring to the correct antecedent, while the incorrect translation contains the pronoun referring to the incorrect antecedent. For the second, the correct translation translates the instance of \textit{it} into the correct disambiguating filler, while the second translation contains the pronoun referring to the incorrect antecedent (see example above).

The sentences for en$\rightarrow$de were common across both the challenge sets developed by \citet{emelin-sennrich-2021-wino}. Hence, the corresponding correct translations from the two challenge sets were used as the ``good'' translation for our evaluation setup. For en$\rightarrow$ru and en$\rightarrow$fr, the source containing the ambiguous pronoun was machine translated and then verified by human annotators to form the ``good'' translation.

\subsection{Untranslated}
\label{sec:untranslated}
MQM defines this error type as ``errors occurring when a text segment that was intended for translation is left untranslated in the target content''. In \textsc{ACES}, we consider both \hyperref[subsec:real_untranslated]{word-level} and \hyperref[subsec:sent-untranslated]{sentence-level} untranslated content.

\subsubsection{Untranslated - Word-Level}
\label{subsec:real_untranslated}

For word-level untranslated content, we manually annotate translations of the FLORES-101 dev and devtest sets: 

\begin{small}
\vspace{0.5cm}
\setlength{\extrarowheight}{0.1cm}
\begin{tabularx}{0.95\columnwidth}{lX}
     SRC (fr): & À l'origine, l'émission mettait en scène des \textbf{comédiens de doublage} amateurs, originaires de l'est du Texas. \\
     REF (de): & Die Sendung hatte ursprünglich lokale Amateur\textbf{synchronsprecher} aus Ost-Texas. \\
     \cmark{} (copy): & Ursprünglich spielte die Show mit Amateur\textbf{synchronsprechern} aus dem Osten von Texas. \\
     \cmark{} (syn.): & Ursprünglich spielte die Show mit Amateur-\textbf{Synchron-Schauspielern} aus dem Osten von Texas. \\
     \xmark: & Ursprünglich spielte die Show mit Amateur-\textbf{Doubling-Schauspielern} aus dem Osten von Texas. \vspace{0.35cm}
\end{tabularx}
\end{small}

We do not only count complete copies as untranslated content but also content that clearly comes from the source language but was only adapted to look more like the target language (as in the example above). If we encounter an untranslated span, we use this translation as the incorrect translation and create a good translation by copying the correct span from the reference and, if possible, a second good translation where we use a synonym for the correct reference span. We manually annotate such untranslated errors for en$\rightarrow$de, fr$\rightarrow$de, de$\rightarrow$en, en$\rightarrow$mr.

\subsubsection{Untranslated - Full Sentences}
\label{subsec:sent-untranslated}
In the case of underperforming machine translation models, sometimes the generated output contains a majority of the tokens from the source language to the extent of copying the entire source sentence.\footnote{Through observations of Swahili $\rightarrow$ English translation; unpublished work} We create a challenge set by simply copying the entire source sentence as the incorrect translation. We used a combination of examples from the FLORES-200, XNLI, and PAWS-X datasets to create these examples. 

We expect that this challenge set is likely to break embedding-based, reference-free evaluation because the representation of the source and the incorrect translation will be the same, thus leading to a higher score.

\subsection{Do Not Translate Errors}
\label{sec:do-not-translate}
This category of errors is defined in MQM as content in the source that should be copied to the output in the source language, but was mistakenly translated into the target language. Common examples of this error type are company names or slogans. Here, we manually create a challenge set based on the PAWS-X data which contains many song titles that should not be translated:

\begin{small}
\vspace{0.5cm}
\setlength{\extrarowheight}{0.1cm}
\begin{tabularx}{0.95\columnwidth}{lX}
     SRC (en): & Dance was one of the inspirations for the exodus - song \textbf{``The Toxic Waltz''}, from their 1989 album ``Fabulous Disaster''. \\
     REF (de): & Dance war eine der Inspirationen für das Exodus-Lied \textbf{„The Toxic Waltz“} von ihrem 1989er Album „Fabulous Disaster“. \\
     \cmark: & Der Tanz war eine der Inspirationen für den Exodus-Song \textbf{„The Toxic Waltz“}, von ihrem 1989er Album „Fabulous Disaster''. \\
     \xmark: & Der Tanz war eine der Inspirationen für den Exodus-Song \textbf{„Der Toxische Walzer“}, von ihrem 1989er Album „Fabulous Disaster''. \vspace{0.35cm}
\end{tabularx}
\end{small}

To construct the challenge set, we use one paraphrase as the good translation and manually translate an English sequence of tokens (e.g. a song title) into German to form the incorrect translation.

\subsection{Overtranslation and Undertranslation}
\label{sec:overtranslation_undertranslation}

Hallucinations from a translation model can often produce a term which is either more generic than the source word or more specific. Within the MQM ontology, the former is referred to as undertranslation while the latter is referred to as overtranslation. 
For example, ``car'' may be substituted with ``vehicle'' (undertranslation) or ``BMW'' (overtranslation). To automate the generation of such errors, we use  Wordnet \citep{miller-1994-wordnet}. In our setup a randomly selected noun from the reference translation is replaced by its corresponding hypernym or hyponym to simulate undertranslation or overtranslation errors, respectively:

\begin{small}
\vspace{0.5cm}
\setlength{\extrarowheight}{0.1cm}
\begin{tabularx}{0.95\columnwidth}{lX}
     SRC (de): & Bob und Ted waren Brüder. Ted ist der \textbf{Sohn} von John. \\
     REF (en): & Bob and Ted were brothers. Ted is John's \textbf{son}. \\
     \cmark: & Bob and Ted were brothers, and Ted is John's \textbf{son}. \\
     \xmark: & Bob and Ted were brothers. Ted is John 's \textbf{male offspring}. \vspace{0.35cm}
\end{tabularx}
\end{small}

During the implementation, we only replaced the first sense listed in Wordnet for the corresponding noun, which may not be appropriate in the given translation. We constructed this challenge set for hypernyms and hyponyms using the PAWS-X dataset, only considering the language pairs where the target language is English.

\subsection{Real-world Knowledge}
\label{sec:real-world-knowledge}
We manually constructed examples each for en$\rightarrow$de and de$\rightarrow$en for the first four phenomena described in this section. We used German-English examples from XNLI, plus English translations from XTREME as the basis for our examples. Typically, we select a single sentence, either the premise or hypothesis from XNLI, and manipulate the MT translations.

\subsubsection{Real-world Knowledge - Textual Entailment}
\label{sec:real_world_knowledge_entailment}
We test whether the metrics can recognise textual entailment -- that is, whether a metric can recognise that the meaning of the source/reference is entailed  by the ``good'' translation.
We construct examples for which the good translation entails the meaning of the original sentence (and its reference). For example, we use the entailment \textit{was murdered} $\rightarrow$ \textit{died} (i.e. if a person is murdered then they must have died) to construct the good translation in the example above. We construct the incorrect translation by replacing the entailed predicate (\textit{died}) with a related but non-entailed predicate (here \textit{was attacked}) -- a person may have been murdered without being attacked, i.e. by being poisoned for example. When constructing our examples we focus solely on leveraging \textit{directional entailments}. We specifically exclude paraphrases as these are bidirectional.

In cases where an antonymous predicate is available, we use that predicate in the incorrect translation. For example, if ``lost'' is in the source/reference, we use ``won'' in the incorrect translation (lost $\not\rightarrow$ won).

\begin{small}
\vspace{0.5cm}
\setlength{\extrarowheight}{0.1cm}
\begin{tabular}{ll}
     SRC (de): & Ein Mann \textbf{wurde ermordet}.\\
     REF (en): & A man \textbf{was murdered}.\\
     \cmark: & A man \textbf{died}.\\
     \xmark{} (omit): & A man \textbf{was attacked}.
     \vspace{0.35cm}
\end{tabular}
\end{small}

\subsubsection{Real-world Knowledge - Hypernyms and Hyponyms}
\label{sec:hypernym-hyponym}

We consider a translation that contains a \textit{hypernym} of a word to be better than one that contains a \textit{hyponym}. For example, whilst translating ``Hund'' (``dog'') with the broader term ``animal'' results in some loss of information, this is preferable over hallucinating information by using a more specific term such as ``labrador'' (i.e. an instance of the hyponym class ``dog''):

\begin{small}
\vspace{0.5cm}
\setlength{\extrarowheight}{0.1cm}
\begin{tabular}{ll}
     SRC (de): & ..., dass der \textbf{Hund} meiner Schwester\\
     & gehört.\\
     REF (en): & ... the \textbf{dog} belonged to my sister.\\
     \cmark{} (hypernym): & ... the \textbf{pet} belonged to my sister.\\
     \xmark{} (hyponym): & ... the \textbf{labrador} belonged to my\\  &sister.
     \vspace{0.35cm}
\end{tabular}
\end{small}

We used Wordnet and WordRel.com\footnote{\url{https://wordrel.com/}} (an online dictionary of words’ relations) to identify hypernyms and hyponyms of nouns within the reference sentences, and used these as substitutions in the MT output: hypernyms are used in the ``good'' translations and hyponyms in the ``incorrect'' translations.

\subsubsection{Real-world Knowledge - Hypernyms and Distractors}
Similar to the \hyperref[sec:hypernym-hyponym]{hypernym vs. hyponym} examples, we construct examples in which the good translation contains a hypernym (here ``pet'') of the word in the reference (here ``dog''). We form the incorrect translation by replacing the original word in the source/reference with a different member from the same class (here ``cat''; both cats and dogs belong to the class of pets). For example:

\begin{small}
\vspace{0.5cm}
\setlength{\extrarowheight}{0.1cm}
\begin{tabular}{ll}
     SRC (de): & ..., dass der \textbf{Hund} meiner Schwester\\
     & gehört.\\
     REF (en): & ... the \textbf{dog} belonged to my sister.\\
     \cmark{} (hypernym): & ... the \textbf{pet} belonged to my sister.\\
     \xmark{} (hyponym): & ... the \textbf{cat} belonged to my sister.
     \vspace{0.35cm}
\end{tabular}
\end{small}

As before, we used Wordnet and WordRel.com to identify hypernyms of nouns present in the reference translation.

\subsubsection{Real-world Knowledge - Antonyms}
\label{sec:antonym}
Similar to the generation of \hyperref[sec:overtranslation_undertranslation]{over- and undertranslations}, we also constructed ``incorrect'' translations by replacing words with their corresponding antonyms from Wordnet. We construct challenge sets for both nouns and verbs.

For nouns, we automatically constructed ``incorrect'' translations by replacing nouns in the reference with their antonyms. The ``good'' translation is not amended. This method may result in noisy replacement of nouns with their respective antonyms. 

In the case of verbs, we manually constructed a more challenging set of examples intended to be used to assess whether the metrics are able to distinguish between translations that contain a synonym versus an antonym of a given word. We replaced verbs in the reference with a synonym to produce the good translation, and with their antonym to produce the incorrect translation:

\begin{small}
\vspace{0.5cm}
\setlength{\extrarowheight}{0.1cm}
\begin{tabular}{ll}
     SRC (de): & Ich \textbf{hasste} jedes Stück der Schule!\\
     REF (en): & I \textbf{hated} every bit of school!\\
     \cmark{} (synonym): & I \textbf{loathed} every bit of school!\\
     \xmark{} (antonym): & I \textbf{loved} every bit of school!
     \vspace{0.35cm}
\end{tabular}
\end{small}

For the verbs challenge set, we consider a translation that contains a synonym of a word in the reference to be a ``good'' translation, and one that contains an antonym of that word to be ``incorrect''. As in the example above the use of synonyms preserves the meaning of the original sentence, and the antonyms introduce a polar opposite meaning.

\subsubsection{Real-world Knowledge - Commonsense}
\label{subsec:real-world-commonsense}

We are also interested in whether evaluation metrics prefer translations that adhere to common sense. To test this, we remove explanatory subordinate clauses from the sources and references in the dataset described in Section~\ref{sec:commonsense-coref}. This guarantees that when choosing between the good and incorrect translation, the metric cannot infer the correct answer from looking at the source or the reference:

\begin{small}
\vspace{0.5cm}
\setlength{\extrarowheight}{0.1cm}
\begin{tabularx}{0.95\columnwidth}{lX}
     SRC (en): & Die Luft im Haus war kühler als in der Wohnung. \\
     REF (de): & The air in the house was cooler than in the apartment. \\
     \cmark: & The air in the house was cooler than in the apartment because \textbf{the apartment} had a broken air conditioner. \\
     \xmark: & The air in the house was cooler than in the apartment because \textbf{the house} had a broken air conditioner. \vspace{0.35cm}
\end{tabularx}
\end{small}

We remove the explanatory subordinate clauses using a sequence of regular expressions. We then pair the shortened source and reference sentences with the full translation that follows commonsense as the good translation and the full translation with the other noun as the incorrect translation. 

Since we present several challenge sets in Section~\ref{sec:source-disambig} where the good translation can only be identified by looking at the source sentence, we also create a version of this challenge set where the explanatory subordinate clause is only removed from the reference but not from the source. By comparing this setup with the results from the setup described above, we achieve another way of quantifying how much a metric considers the source.

\subsection{Wrong Language}
\label{sec:wrong_language}
Most of the representations obtained from large multilingual language models do not explicitly use the language identifier (id) as an input while encoding a sentence. Here, we are interested in checking whether sentences which have similar meanings are closer together in the representation space of neural MT evaluation metrics, irrespective of their language.  We create a challenge set for embedding-based metrics where the incorrect translation is in a similar language (same typology/same script) to the reference (e.g. a Catalan translation may be used as the incorrect translation if the target language is Spanish). Note that this is also a common error with multilingual machine translation models. We constructed these examples using the FLORES-200 dataset where the ``good'' translation was the automatic translation and the ``incorrect'' translation was the reference from a language similar to the target language:

\begin{small}
\vspace{0.5cm}
\setlength{\extrarowheight}{0.1cm}
\begin{tabularx}{0.95\columnwidth}{lX}
     SRC (en): & Cell comes from the Latin word cella which means small room. \\
     REF (es): & El término célula deriva de la palabra latina cella, que quiere decir «cuarto pequeño». \\
     \cmark\ (es): & La célula viene de la palabra latina cella que significa habitación pequeña. \\
     \xmark\ (ca): & Cèl·lula ve de la paraula llatina cella, que vol dir habitació petita. \vspace{0.35cm}
\end{tabularx}
\end{small}

We construct two categories within this challenge set: one where the target language is a higher-resource language and the incorrect language is a lower-resource language and vice-versa. The languages we consider are (\texttt{src-tgt-sim}): en-hi-mr, en-es-ca, en-cs-pl, fr-mr-hi,  en-pl-cs, and en-ca-es.

Note that if we were to compare references for different languages and not an automatic translation vs. a reference, this challenge set should be considered unsolvable for reference-free metrics if there is no way to specify the desired target language. But in this case, we expect reference-free metrics to prefer the reference that we use as the ``incorrect translation'' since there may be translation errors in the automatically translated ``good translation''.

\subsection{Fluency}
Although the focus of \textsc{ACES} is on accuracy errors, we also include a small set of fluency errors for the punctuation category. Future work might consider expanding this set to include other categories of fluency errors. 

\subsubsection{Punctuation}
\label{sec:punctuation}
We assess the effect of deleting and substituting punctuation characters. We employ four strategies: 1) deleting all punctuation, 2) deleting only quotation marks (i.e. removing indications of quoted speech), 3) deleting only commas (i.e. removing clause boundary markers), 4) replacing exclamation points with question marks (i.e. statement $\rightarrow$ question).

In strategies 1 and, especially, 3 and 4, some of the examples may also contain accuracy-related errors. For example, the meaning of the sentence could be changed in the incorrect translation if we remove a comma, e.g. in the (in)famous example ``Let's eat, Grandma!'' vs. ``Let's eat Grandma!''. We use the TED Talks from the WMT 2018 English-German pronoun translation evaluation test suite and apply all deletions and substitutions automatically.

\section{Evaluation Methodology}
\label{sec:eval_methodology}

We shall now briefly describe the metrics that participated in the challenge set shared task. The organisers of the shared task also provided scores by a number of baseline metrics, as described below.

\subsection{Baseline Metrics}
\textbf{BLEU} \citep{papineni-etal-2002-bleu} compares the token-level n-grams of the hypothesis with the reference translation and then computes a precision score weighted by a brevity penalty.\\

\noindent\textbf{spBLEU} \citep{goyal-etal-2022-flores} is BLEU computed over text tokenised with a single language-agnostic SentencePiece subword model. The spBLEU baselines, \textsc{f101spBLEU} and \textsc{f200spBLEU}, are named according to whether the SentencePiece tokeniser \citep{kudo-richardson-2018-sentencepiece} was trained using data from the FLORES-101 or FLORES-200 languages.\\

\noindent \textbf{chrF} \citep{popovic-2017-chrf} evaluates translation outputs based on a character n-gram F-score by computing overlaps between the hypothesis and the reference.\\

\noindent \textbf{BERTScore} \citep{DBLP:conf/iclr/ZhangKWWA20} uses contextual embeddings from pre-trained language models to compute the similarity between the tokens in the reference and the generated translation using cosine similarity. The similarity matrix is used to compute precision, recall, and F1-scores.\\

\noindent \textbf{BLEURT20} \citep{sellam-etal-2020-learning} is a BERT-based \citep{devlin-etal-2019-bert} regression model, which is first trained on  scores of automatic metrics/similarity of pairs of reference sentences and their corrupted counterparts. It is then fine-tuned on the WMT human evaluation data to produce a score for a hypothesis given a reference translation.\\

\noindent \textbf{COMET-20}  \citep{rei-etal-2020-comet} uses a cross-lingual encoder (XLM-R \citep{conneau-etal-2020-unsupervised}) and pooling operations to obtain sentence-level representations of the source, hypothesis, and reference. These sentence embeddings are combined and then passed through a feedforward network to produce a score. \textsc{COMET} is trained on human evaluation scores of machine translation systems submitted to WMT until 2020. \\

\noindent \textbf{COMET-QE} was trained similarly to \textsc{COMET-20} but as this is a reference-free metric, only the source and the hypothesis are combined to produce a final score. \\

\noindent \textbf{YiSi-1} \citep{lo-2019-yisi}
measures the semantic similarity between the hypothesis and the reference by using cosine similarity scores of multilingual representations at the lexical level. It optionally uses a semantic role labeller to obtain structural similarity. Finally, a weighted f-score based on structural and lexical similarity is used for scoring the hypothesis against the reference.

\subsection{Metrics Submitted to WMT 2022}
We list the descriptions provided by the authors of the respective metrics and refer the reader to the relevant system description papers for further details. \\

\noindent \textbf{COMET-22} \citep{COMET:WMT22} is an ensemble between a vanilla \textsc{COMET} model trained with Direct Assessment (DA) scores  and a Multitask model that is trained on regression (MQM regression) and sequence tagging (OK/BAD word identification from MQM span annotations). These models are ensembled together using a hyperparameter search that weights different features extracted from these two evaluation models and combines them into a single score.
The vanilla \textsc{COMET} model is trained with DA’s ranging 2017 to 2020 while the Multitask model is trained using DA’s ranging from 2017 to 2020 plus MQM annotations from 2020 (except for en-ru that uses TedTalk annotations from 2021).\\

\noindent \textbf{Metric-X} is a massive multi-task metric, which fine tunes large language model checkpoints such as mT5 on a variety of human feedback data such as Direct Assessment, MQM, QE, NLI and Summarization Eval. Scaling up the metric is the key to unlocking quality and makes the model work in difficult settings such as evaluating without a reference, evaluating short queries, distinguishing high quality outputs, and evaluating on other generation tasks such as summarisation. The four metrics are referred to according to the mT5 model variant used (xl or xxl) and the fine-tuning data: \textsc{metricx\_*\_DA\_2019} only used 2015-19 Direct Assessment data for fine-tuning, whereas \textsc{metricx\_*\_MQM\_2020} used a mixture of Direct Assessment 2015-19 and MQM 2020 data.

\noindent \textbf{MS-COMET-22} and \textbf{MS-COMET-QE-22} \citep{MS-COMET:WMT22} are built on top of the COMET \citep{rei-etal-2020-comet} architecture. They are trained on a several times larger set of human judgements covering 113 languages and covering 15 domains. Furthermore, the authors propose filtering of human judgements with potentially low quality. \textsc{MS-COMET-22} receives the source, the MT hypothesis and the human reference as input, while  \textsc{MS-COMET-QE} calculates scores in a quality estimation fashion with access only to the source segment and the MT hypothesis.\\

\noindent \textbf{UniTE} \citep{UNITE:WMT22}, Unified Translation Evaluation, is a metric approach where the model-based metrics can possess the ability of evaluating translation outputs following all three evaluation scenarios, i.e. source-only, reference-only, and source-reference-combined. These are referred to in this paper as \textsc{UniTE-src}, \textsc{UniTE-ref}, and \textsc{UniTE} respectively. \\

\noindent \textbf{COMET-Kiwi} \citep{COMET:WMT22} ensembles two QE models similarly to \textsc{COMET-22}. The first model follows the classic Predictor-Estimator QE architecture where MT and source are encoded together. This model is trained on DAs ranging 2017 to 2019 and then fine-tuned on DAs from MLQE-PE (the official DA from the QE shared task). The second model is the same multitask model used in the \textsc{COMET-22} submission but without access to a reference translation. This means that this model is a multitask model trained on regression and sequence tagging. Both models are ensembled together using a hyperparameter search that weights different features extracted from these two QE models and combines them into a single score.\\

\noindent Huawei submitted several metrics to the shared task \citep{HWTSC-Metrics:WMT22}. \textbf{Cross-QE} is a submission based on the \textsc{COMET-QE} architecture.  \textbf{HWTSC-Teacher-Sim} is a reference-free metric constructed by fine-tuning the multilingual Sentence BERT model: paraphrase-multilingual-mpnet-base-v2 \citep{reimers-gurevych-2019-sentence}. \textbf{HWTSC-TLM} is a reference-free metric which only uses a target-side language model and only uses the system translations as input. \textbf{KG-BERTScore} is a reference-free machine translation evaluation metric, which incorporates a multilingual knowledge graph into BERTScore by linearly combining the results of BERTScore and bilingual named entity matching. \\

\noindent \textbf{MATESE} metrics \citep{MATESE:WMT22} leverage Transformer-based multilingual encoders to identify error spans in translations, and classify their severity between MINOR and MAJOR. The quality score returned for a translation is computed following the MQM error weighting introduced in  \citet{freitag-etal-2021-experts}. \textsc{MATESE} is reference-based, while \textbf{MATESE-QE} is its reference-free version, with the source sentence used in place of the reference.\\

\noindent \textbf{MEE} \citep{MEE2020} is an automatic evaluation metric that leverages the similarity between embeddings of words in candidate and reference sentences to assess translation quality, focusing mainly on adequacy. Unigrams are matched based on their surface forms, root forms and meanings which aims to capture lexical, morphological and semantic equivalence. Semantic evaluation is achieved by using pretrained fasttext embeddings provided by Facebook to calculate the word similarity score between the candidate and reference words. MEE computes an evaluation score using three modules namely exact match, root match and synonym match. In each module, fmean-score is calculated using the harmonic mean of precision and recall by assigning more weightage to recall. The final translation score is obtained by taking average of fmean-scores from individual modules.\\

\noindent \textbf{MEE2} and \textbf{MEE4} \citep{MEE:WMT22} are improved versions of \textsc{MEE}, focusing on computing contextual and syntactic equivalences along with lexical, morphological and semantic similarity. The intent is to capture fluency and context of the MT outputs along with their adequacy. Fluency is captured using syntactic similarity and context is captured using sentence similarity leveraging sentence embeddings. The final sentence translation score is the weighted combination of three similarity scores: a) Syntactic Similarity achieved by modified BLEU score; b) Lexical, Morphological and Semantic Similarity: measured by explicit unigram matching similar to MEE score; c) Contextual Similarity: Sentence similarity scores are calculated by leveraging sentence embeddings of Language-Agnostic BERT models.\\

\noindent \textbf{REUSE} \citep{REUSE:WMT22} is a REference-free UnSupervised quality Estimation Metric. This is a bilingual untrained metric. It estimates the translation quality at chunk-level and sentence-level. Source and target sentence chunks are retrieved by using a multi-lingual chunker. Chunk-level similarity is computed by leveraging BERT contextual word embeddings and sentence similarity scores are calculated by leveraging sentence embeddings of Language-Agnostic BERT models. The final quality estimation score is obtained by mean pooling the chunk-level and sentence-level similarity scores.

\subsection{Evaluation of Metrics}

For all phenomena in \textsc{ACES} where we generated more than 1,000 examples, we randomly subsample 1,000 examples according to the per language pair distribution to include in the final challenge set to keep the evaluation of new metrics tractable.

We follow the evaluation of the challenge sets from the 2021 edition of the WMT metrics shared task \citep{freitag-etal-2021-results} and report performance with Kendall's tau-like correlation. This metric measures the number of times a metric scores the good translation above the incorrect translation (concordant) and equal to or lower than the incorrect translation (discordant):
\begin{center}
\begin{align*}
    \tau = \frac{concordant - discordant}{concordant + discordant}\\
\end{align*}
\end{center}
Ties are considered as discordant. Note that a higher $\tau$ indicates a better performance and that the values can range between -1 and 1. 

\section{Results}
\label{sec: Results}

\subsection{Phenomena-level Results}
We start by providing a broad overview of metric performance on the different categories of phenomena. We compute Kendall's tau-like correlation scores (Section~\ref{sec:eval_methodology}) for the 24 metrics which a) provide segment-level scores and b) provide scores for all language pairs and directions in \textsc{ACES}. We first compute the correlation scores for all of the individual phenomena and then take the average score over all phenomena in each of the nine top-level accuracy categories in \textsc{ACES} plus the fluency category \hyperref[sec:punctuation]{punctuation} (see Table~\ref{tab:analysis_overview}).

The performance of the metrics varies greatly and there is no clear \textit{winner} in terms of performance across all of the categories. There is also a high degree of variation in terms of metric performance when each category is considered in isolation. Whilst each of the categories proves challenging for at least one metric, some categories are more challenging than others. For example, looking at the average scores in the last row of Table~\ref{tab:analysis_overview}, and without taking outliers into account, we might conclude that \hyperref[sec:addition-omission]{addition}, \hyperref[sec:overtranslation_undertranslation]{undertranslation}, \hyperref[sec:real-world-knowledge]{real-world knowledge}, and \hyperref[sec:wrong_language]{wrong language} (all with average Kendall tau-like correlation of $<$ 0.3) present more of a challenge than the other categories. On the other hand, for \hyperref[sec:addition-omission]{omission} and \hyperref[sec:do-not-translate]{do not translate} (with an average Kendall tau-like correlation of $>$ 0.7) metric performance is generally rather high. 

We also observe variation in terms of the performance of metrics belonging to the baseline, reference-based, and reference-free groups. For example, the baseline metrics appear to struggle more on the \hyperref[sec:overtranslation_undertranslation]{overtranslation and undertranslation} categories than the metrics belonging to the other groups. Reference-based metrics also appear to perform better overall on the \hyperref[sec:untranslated]{untranslated} category than the reference-free metrics. This makes sense as a comparison with the reference is likely to highlight tokens that ought to have been translated.
\begin{sidewaystable*}[ht]
\small
\setlength{\tabcolsep}{3.75pt}
\centering
\begin{tabular}{@{}lccccccccccc@{}}
\toprule
 & \hyperref[sec:addition-omission]{\textbf{addition}} & \hyperref[sec:addition-omission]{\textbf{omission}} & \hyperref[sec:source-disambig]{\textbf{mistranslation}} & \hyperref[sec:untranslated]{\textbf{untranslated}} & \hyperref[sec:do-not-translate]{\textbf{do not}} & \hyperref[sec:overtranslation_undertranslation]{\textbf{overtranslation}} & \hyperref[sec:overtranslation_undertranslation]{\textbf{undertranslation}} & \hyperref[sec:real-world-knowledge]{\textbf{real-world}} & \hyperref[sec:wrong_language]{\textbf{wrong}} & \hyperref[sec:punctuation]{\textbf{punctuation}} & \textbf{ACES-}\\ 
&  &  &  &  & \hyperref[sec:do-not-translate]{\textbf{translate}} &  &  & \hyperref[sec:real-world-knowledge]{\textbf{knowledge}} & \hyperref[sec:wrong_language]{\textbf{language}} &  & \textbf{Score}\\ 
\midrule
\textit{\textbf{Examples}}     & \textit{999}                 & \textit{999}                 & \textit{24457}                     & \textit{1300}                    & \textit{100}                         & \textit{1000}                       & \textit{1000}                        & \textit{2948}                            & \textit{2000}                      & \textit{1673}                   \\ 
\midrule
BLEU                    & \phantom{-}0.748    & \phantom{-}0.435    & -0.229         & \phantom{-}0.353        & \phantom{-}0.600            & -0.838          & -0.856           & -0.768               & \phantom{-}0.661          & \phantom{-}0.638       & -2.79      \\
f101spBLEU              & \phantom{-}0.662    & \phantom{-}0.590    & -0.084         & \phantom{-}0.660        & \phantom{-}0.940            & -0.738          & -0.826           & -0.405               & \phantom{-}0.638          & \phantom{-}0.639       & -0.09      \\
f200spBLEU              & \phantom{-}0.664    & \phantom{-}0.590    & -0.082         & \phantom{-}0.687        & \phantom{-}0.920            & -0.752          & -0.794           & -0.394               & \phantom{-}0.658          & \phantom{-}0.648       & \phantom{-}0.06       \\
chrF                    & \phantom{-}0.642    & \phantom{-}0.784    & \phantom{-}0.162          & \colorbox[HTML]{B2EAB1}{\textbf{\phantom{-}0.781}}        & \colorbox[HTML]{B2EAB1}{\textbf{\phantom{-}0.960}}            & -0.696          & -0.592           & -0.294               & \colorbox[HTML]{B2EAB1}{\textbf{\phantom{-}0.691}}          & \phantom{-}0.743       & \phantom{1}3.71       \\
BERTScore               & \colorbox[HTML]{B2EAB1}{\textbf{\phantom{-}0.880}}    & \phantom{-}0.750    & \phantom{-}0.320          & \phantom{-}0.767        & \colorbox[HTML]{B2EAB1}{\textbf{\phantom{-}0.960}}            & -0.110          & -0.190           & \phantom{-}0.031                & \phantom{-}0.563          & \colorbox[HTML]{B2EAB1}{\textbf{\phantom{-}0.849}}       & 10.65      \\
BLEURT-20               & \phantom{-}0.437    & \phantom{-}0.810    & \phantom{-}0.429          & \phantom{-}0.748        & \phantom{-}0.860            & \phantom{-}0.200           & \phantom{-}0.014            & \phantom{-}0.401                & \phantom{-}0.533          & \phantom{-}0.649       & 12.06      \\
COMET-20                & \phantom{-}0.437    & \phantom{-}0.808    & \phantom{-}0.378          & \phantom{-}0.748        & \phantom{-}0.900            & \phantom{-}0.314           & \phantom{-}0.112            & \phantom{-}0.267                & \phantom{-}0.033          & \phantom{-}0.706       & 12.27      \\
COMET-QE                & -0.538   & \phantom{-}0.397    & \phantom{-}0.378          & \phantom{-}0.135        & \phantom{-}0.120            & \phantom{-}0.622           & \phantom{-}0.442            & \phantom{-}0.322                & -0.505         & \phantom{-}0.251       & \phantom{1}6.61       \\
YiSi-1                  & \phantom{-}0.770    & \phantom{-}0.866    & \phantom{-}0.356          & \phantom{-}0.730        & \phantom{-}0.920            & -0.062          & -0.076           & \phantom{-}0.110                & \phantom{-}0.431          & \phantom{-}0.734       & 11.53      \\
\midrule
COMET-22                & \phantom{-}0.333    & \phantom{-}0.806    & \phantom{-}0.566          & \phantom{-}0.536        & \phantom{-}0.900            & \phantom{-}0.690           & \phantom{-}0.538            & \phantom{-}0.574                & -0.318         & \phantom{-}0.539       & 16.41      \\
metricx\_xl\_DA\_2019   & \phantom{-}0.395    & \phantom{-}0.852    & \phantom{-}0.545          & \phantom{-}0.722        & \phantom{-}0.940            & \phantom{-}0.692           & \phantom{-}0.376            & \colorbox[HTML]{B2EAB1}{\textbf{\phantom{-}0.740}}                & \phantom{-}0.521          & \phantom{-}0.670       & 17.29      \\
metricx\_xl\_MQM\_2020  & -0.281   & \phantom{-}0.670    & \phantom{-}0.523          & \phantom{-}0.579        & {-}0.740            & \phantom{-}0.718           & \colorbox[HTML]{B2EAB1}{\textbf{\phantom{-}0.602}}            & \phantom{-}0.705                & -0.126         & \phantom{-}0.445       & 13.10      \\
metricx\_xxl\_DA\_2019  & \phantom{-}0.303    & \phantom{-}0.832    & \phantom{-}0.580          & \phantom{-}0.762        & \phantom{-}0.920            & \phantom{-}0.572           & \phantom{-}0.246            & \phantom{-}0.691                & \phantom{-}0.250          & \phantom{-}0.630       & 15.35      \\
metricx\_xxl\_MQM\_2020 & -0.099   & \phantom{-}0.534    & \phantom{-}0.578          & \phantom{-}0.651        & \phantom{-}0.880            & \colorbox[HTML]{B2EAB1}{\textbf{\phantom{-}0.752}}           & \phantom{-}0.552            & \phantom{-}0.712                & -0.321         & \phantom{-}0.369       & 13.54      \\
MS-COMET-22             & -0.219   & \phantom{-}0.686    & \phantom{-}0.397          & \phantom{-}0.504        & \phantom{-}0.700            & \phantom{-}0.548           & \phantom{-}0.290            & \phantom{-}0.230                & \phantom{-}0.041          & \phantom{-}0.508       & 10.03      \\
UniTE                   & \phantom{-}0.439    & \phantom{-}0.876    & \phantom{-}0.501          & \phantom{-}0.571        & \phantom{-}0.920            & \phantom{-}0.496           & \phantom{-}0.302            & \phantom{-}0.624                & -0.337         & \phantom{-}0.793       & 14.93      \\
UniTE-ref               & \phantom{-}0.359    & \phantom{-}0.868    & \phantom{-}0.535          & \phantom{-}0.412        & \phantom{-}0.840            & \phantom{-}0.640           & \phantom{-}0.398            & \phantom{-}0.585                & -0.387         & \phantom{-}0.709       & 15.52      \\
\midrule
COMETKiwi               & \phantom{-}0.361    & \phantom{-}0.830    & \colorbox[HTML]{B2EAB1}{\textbf{\phantom{-}0.631}}          & \phantom{-}0.230        & \phantom{-}0.780            & \phantom{-}0.738           & \phantom{-}0.574            & \phantom{-}0.582                & -0.359         & \phantom{-}0.490       & 16.95      \\
Cross-QE                & \phantom{-}0.163    & \phantom{-}0.876    & \phantom{-}0.546          & -0.094       & \phantom{-}0.320            & \phantom{-}0.726           & \phantom{-}0.506            & \phantom{-}0.446                & -0.374         & \phantom{-}0.455       & 14.43      \\
HWTSC-Teacher-Sim       & -0.031   & \phantom{-}0.495    & \phantom{-}0.406          & -0.269       & \phantom{-}0.700            & \phantom{-}0.552           & \phantom{-}0.456            & \phantom{-}0.261                & -0.021         & \phantom{-}0.271       & 10.09      \\
HWTSC-TLM               & -0.363   & \phantom{-}0.345    & \phantom{-}0.384          & \phantom{-}0.154        & -0.040           & \phantom{-}0.544           & \phantom{-}0.474            & \phantom{-}0.071                & -0.168         & \phantom{-}0.634       & \phantom{1}7.00       \\
KG-BERTScore            & \phantom{-}0.790    & \phantom{-}0.812    & \phantom{-}0.489          & -0.456       & \phantom{-}0.760            & \phantom{-}0.654           & \phantom{-}0.528            & \phantom{-}0.487                & \phantom{-}0.306          & \phantom{-}0.255       & \colorbox[HTML]{B2EAB1}{\textbf{17.49}}      \\
MS-COMET-QE-22          & -0.177   & \phantom{-}0.678    & \phantom{-}0.439          & \phantom{-}0.388        & \phantom{-}0.240            & \phantom{-}0.518           & \phantom{-}0.386            & \phantom{-}0.248                & -0.197         & \phantom{-}0.523       & \phantom{1}9.95       \\
UniTE-src               & \phantom{-}0.285    & \colorbox[HTML]{B2EAB1}{\textbf{\phantom{-}0.930}}    & \phantom{-}0.599          & -0.615       & \phantom{-}0.860            & \phantom{-}0.698           & \phantom{-}0.540            & \phantom{-}0.537                & -0.417         & \phantom{-}0.733       & 15.70      \\
\midrule
Average                 & \phantom{-}0.290    & \phantom{-}0.713    & \phantom{-}0.389          & \phantom{-}0.404        & \phantom{-}0.735            & \phantom{-}0.312           & \phantom{-}0.167            & \phantom{-}0.282                & \phantom{-}0.075          & \phantom{-}0.578       & 10.91  \\   
\bottomrule
\end{tabular}
\caption{Average Kendall’s tau-like correlation results for the nine top level categories in the \textsc{ACES} ontology, plus the additional fluency category: punctuation.  The horizontal lines delimit baseline metrics (top), participating reference-based metrics (middle) and participating reference-free metrics (bottom). The best result for each category is denoted by bold text with a green highlight. Note that \textit{Average} is an average over averages. The last column shows the \textsc{ACES}-Score, a weighted sum of the correlations. The \textsc{ACES}-Score ranges from -29.1 (all phenomena have a correlation of -1) to 29.1 (all phenomena have a correlation of +1).}
\label{tab:analysis_overview}
\end{sidewaystable*}
\afterpage{\clearpage}

\subsection{ACES Score}

To analyse general, high-level, performance trends of the metrics on the \textsc{ACES} challenge set, we define a weighted combination of the top-level categories to derive a single score. We call this score the ``ACES - Score'':

\begin{equation}
\small
\label{eq:aces-score}
\textsc{ACES} = sum \left\{
\begin{aligned}
    \quad 5 * \tau_{\text{addition}}\\
   \quad  5 * \tau_{\text{omission}}\\
   \quad  5 * \tau_{\text{mistranslation}}\\
   \quad  1 * \tau_{\text{untranslated}}\\
   \quad  1 * \tau_{\text{do not translate}}\\
   \quad  5 * \tau_{\text{overtranslation}}\\
   \quad  5 * \tau_{\text{undertranslation}}\\
   \quad  1 * \tau_{\text{real-world knowledge}}\\
   \quad  1 * \tau_{\text{wrong language}}\\
   \quad  0.1 * \tau_{\text{punctuation}}\\
\end{aligned}
\right\} \vspace{0.5cm}
\end{equation}

The weights correspond to the values under the MQM framework that \citet{freitag-etal-2021-experts} recommend for major (weight$=$5), minor (weight$=$1) and fluency/punctuation errors (weight$=$0.1). We determined that \hyperref[sec:untranslated]{untranslated}, \hyperref[sec:do-not-translate]{do not translate} and \hyperref[sec:wrong_language]{wrong language} errors should be counted as minor errors because they can be identified automatically with language detection tools and should also be easy to spot in post-editing. We also include \hyperref[sec:real-world-knowledge]{real-world knowledge} under minor errors since we do not expect that current MT evaluation metrics have any notion of real-world knowledge and we do not want to punish them too severely if they do not perform well on this challenge set.

We caution that our weighting for the \textsc{ACES}-Score is not ideal, as some phenomena within a broad category might be more difficult than others. Still, we believe that an \textsc{ACES}-Score will be helpful to quickly identify changes in performance of a metric (e.g. following modifications), prior to conducting in-depth analyses at the category and sub-category levels. The \textsc{ACES}-Score ranges from -29.1 (all phenomena have a correlation of -1) to 29.1 (all phenomena have a correlation of +1).

The \textsc{ACES}-Score results can be seen in the last column of Table~\ref{tab:analysis_overview}. Using the \textsc{ACES}-Score, we can see at a glance that the majority of the metrics submitted to the WMT 2022 shared task outperform the baseline metrics. Interestingly, many reference-free metrics also perform on par with reference-based metrics. The best performing metric is a reference-free metric, namely \textsc{KG-BERTScore}, closely followed by the reference-based metric \textsc{metricx\_xl\_DA\_2019}. Perhaps unsurprisingly, the worst performing metric is \textsc{BLEU}. However, we caution against making strong claims about which metrics perform \textit{best} or \textit{worst} on the challenge set based on this score alone. Instead, we recommend that \textsc{ACES} be used to highlight general trends as to what the outstanding issues are for MT evaluation metrics. More fine-grained analyses are reported in the following sections.

More generally, work on analysing system performance on \textit{\textsc{ACES}} prompts the question: What is the definition of a good metric? One might consider that a \textit{good} metric exhibits a strong correlation with human judgements on whether a translation is good/bad \textit{and}  assigns sufficiently different scores to a good vs. an incorrect translation. The latter criterion would provide evidence of the ability of the metric to discriminate reliably between good and incorrect translations, but it may be difficult to establish what this difference should be, especially without knowing to what degree the translations are good/bad without human judgements and because the scales of different metrics are not comparable. We leave an analysis of metrics' confidence on different error types for future work.

\subsection{Mistranslation Results}
Next, we drill down to the fine-grained categories of the largest category: \textit{mistranslation}. We present metric performance on its sub-level categories in Table~\ref{tab:analysis_mistranslation}. Again, we find that performance on the different sub-categories is variable, with no clear \textit{winner} among the metrics. The results suggest that \hyperref[sec:hallucination]{hallucination} phenomena are generally more challenging than \hyperref[sec:discourse]{discourse-level} phenomena. Performance on the \hyperref[sec:hallucination]{hallucination} sub-category is poor overall, although it appears to be particularly challenging for the baseline metrics. We present additional, more fine-grained, performance analyses for individual phenomena in Section \ref{sec:analysis}. 
\begin{table}[t]
\centering
\small
\setlength{\tabcolsep}{5pt}
\setlength{\fboxsep}{0.5pt} % Reduces the size of the colorbox cell shading
\begin{tabular}{@{}lccc@{}}
\toprule
                    & \hyperref[sec:discourse]{\textbf{disco.}} & \hyperref[sec:hallucination]{\textbf{halluci.}} & \textbf{other}         \\ \midrule
\textit{\textbf{Examples}}          & \textit{3698}   & \textit{10270} & \textit{10489} \\ \midrule
BLEU                    & -0.048                                         & -0.420                                             & -0.251                                     \\
f101spBLEU              & \phantom{-}0.105                                          & -0.206                                             & -0.153                                     \\
f200spBLEU              & \phantom{-}0.094                                          & -0.191                                             & -0.149                                     \\
chrF                    & \phantom{-}0.405                                          & -0.137                                             & \phantom{-}0.161                                      \\
BERTScore               & \phantom{-}0.567                                          & -0.058                                             & \phantom{-}0.362                                      \\
BLEURT-20               & \phantom{-}0.695                                          & \phantom{-}0.142                                              & \phantom{-}0.402                                      \\
COMET-20                & \phantom{-}0.641                                          & \phantom{-}0.016                                              & \phantom{-}0.399                                      \\
COMET-QE                & \phantom{-}0.666                                          & \phantom{-}0.303                                              & \phantom{-}0.208                                      \\
YiSi-1                  & \phantom{-}0.609                                          & \phantom{-}0.019                                              & \phantom{-}0.368                                      \\
\midrule
COMET-22                & \phantom{-}0.682                                          & \phantom{-}0.461                                              & \phantom{-}0.542                                      \\
metricx\_xl\_DA\_2019   & \phantom{-}0.701                                          & \phantom{-}0.493                                              & \phantom{-}0.458                                      \\
metricx\_xl\_MQM\_2020  & \phantom{-}0.573                                          & \phantom{-}0.677                                              & \phantom{-}0.394                                      \\
metricx\_xxl\_DA\_2019  & \phantom{-}0.768                                          & \phantom{-}0.541                                              & \phantom{-}0.463                                      \\
metricx\_xxl\_MQM\_2020 & \phantom{-}0.716                                          & \colorbox[HTML]{B2EAB1}{\textbf{\phantom{-}0.713}}                                              & \phantom{-}0.392                                      \\
MS-COMET-22             & \phantom{-}0.645                                          & \phantom{-}0.148                                              & \phantom{-}0.360                                      \\
UniTE                   & \phantom{-}0.746                                          & \phantom{-}0.322                                              & \phantom{-}0.424                                      \\
UniTE-ref               & \colorbox[HTML]{B2EAB1}{\textbf{\phantom{-}0.776}}                                          & \phantom{-}0.396                                              & \phantom{-}0.437                                      \\
\midrule
COMETKiwi               & \phantom{-}0.733                                          & \phantom{-}0.493                                              & \colorbox[HTML]{B2EAB1}{\textbf{\phantom{-}0.637}}                                      \\
Cross-QE                & \phantom{-}0.644                                          & \phantom{-}0.395                                              & \phantom{-}0.563                                      \\
HWTSC-Teacher-Sim       & \phantom{-}0.594                                          & \phantom{-}0.296                                              & \phantom{-}0.330                                      \\
HWTSC-TLM               & \phantom{-}0.756                                          & \phantom{-}0.306                                              & \phantom{-}0.151                                      \\
KG-BERTScore            & \phantom{-}0.593                                          & \phantom{-}0.387                                              & \phantom{-}0.472                                      \\
MS-COMET-QE-22          & \phantom{-}0.626                                          & \phantom{-}0.243                                              & \phantom{-}0.416                                      \\
UniTE-src               & \phantom{-}0.772                                          & \phantom{-}0.463                                              & \phantom{-}0.551                                      \\
\midrule
Average                 & \phantom{-}0.586                                          & \phantom{-}0.242                                              & \phantom{-}0.331                                     \\
\bottomrule
\end{tabular}
\caption{Average Kendall’s tau-like correlation results for the sub-level categories in mistranslation: \hyperref[sec:discourse]{\textbf{disco}urse-level}, \hyperref[sec:hallucination]{\textbf{halluci}nation}, and \textbf{other} errors.  The horizontal lines delimit baseline metrics (top), participating reference-based metrics (middle) and participating reference-free metrics (bottom). The best result for each category is denoted by bold text with a green highlight. Note that \textit{Average} is an average over averages.}
\label{tab:analysis_mistranslation}
\end{table}

\subsection{Language-level Results}
\begin{table}[ht]
\small
\centering
\setlength{\tabcolsep}{1.5pt}
\setlength{\fboxsep}{0.5pt} % Reduces the size of the colorbox cell shading
\begin{tabular}{@{}lcccc@{}}
\toprule
                    & \textbf{trained} & \textbf{en-x} & \textbf{x-en} & \textbf{x-y}         \\
\midrule
\textit{\textbf{Examples}}          & 8871  & 12695                   & 17966                   & 5815                    \\ \midrule
BLEU                    & \phantom{-}0.009 & \phantom{-}0.225                   & -0.370                  & -0.121                  \\
f101spBLEU              & \phantom{-}0.148  & \phantom{-}0.170                   & -0.290                  & -0.022                  \\
f200spBLEU              & \phantom{-}0.140  & \phantom{-}0.442                   & -0.286                  & -0.004                  \\
chrF                    & \phantom{-}0.325  & \phantom{-}0.392                   & -0.047                  & \phantom{-}0.098                   \\
BERTScore               & \phantom{-}0.479  & \phantom{-}0.031                   & \phantom{-}0.173                   & \phantom{-}0.125                   \\
BLEURT-20               & \phantom{-}0.541  & \phantom{-}0.327                   & \phantom{-}0.280                   & \phantom{-}0.257                   \\
COMET-20                & \phantom{-}0.495  & \phantom{-}0.379                   & \phantom{-}0.278                   & \phantom{-}0.121                   \\
COMET-QE 			    & \phantom{-}0.356  & \phantom{-}0.166                   & \phantom{-}0.144                   & \phantom{-}0.168                   \\
YiSi-1                  & \phantom{-}0.476  & \phantom{-}0.520                   & \phantom{-}0.185                   & \phantom{-}0.150                   \\ \midrule
COMET-22                & \phantom{-}0.599  & \phantom{-}0.486                   & \phantom{-}0.554                   & \phantom{-}0.355                   \\
metricx\_xl\_DA\_2019   & \phantom{-}0.622  & \phantom{-}0.458                   & \phantom{-}0.456                   & \colorbox[HTML]{B2EAB1}{\textbf{\phantom{-}0.551}}                   \\
metricx\_xl\_MQM\_2020  & \phantom{-}0.608  & \phantom{-}0.567                   & \phantom{-}0.452                   & \phantom{-}0.509                   \\
metricx\_xxl\_DA\_2019  & \phantom{-}0.631  & \phantom{-}0.431                   & \phantom{-}0.462                   & \phantom{-}0.528                   \\
metricx\_xxl\_MQM\_2020 & \phantom{-}0.605  & \colorbox[HTML]{B2EAB1}{\textbf{\phantom{-}0.572}}                   & \phantom{-}0.487                   & \phantom{-}0.502                   \\
MS-COMET-22             & \phantom{-}0.415  & \phantom{-}0.312                   & \phantom{-}0.323                   & \phantom{-}0.117                   \\
UniTE                   & \phantom{-}0.635  & \phantom{-}0.452                   & \phantom{-}0.406                   & \phantom{-}0.283                   \\
UniTE-ref               & \phantom{-}0.619  & \phantom{-}0.313                   & \phantom{-}0.413                   & \phantom{-}0.305                   \\ \midrule
COMETKiwi               & \phantom{-}0.620  & \phantom{-}0.510                   & \colorbox[HTML]{B2EAB1}{\textbf{\phantom{-}0.694}}                   & \phantom{-}0.468                   \\
Cross-QE                & \phantom{-}0.598  & \phantom{-}0.401                   & \phantom{-}0.552                   & \phantom{-}0.291                   \\
HWTSC-Teacher-Sim       & \phantom{-}0.497  & \phantom{-}0.357                   & \phantom{-}0.352                   & \phantom{-}0.149                   \\
HWTSC-TLM               & \phantom{-}0.538  & \phantom{-}0.519                   & \phantom{-}0.167                   & \phantom{-}0.194                   \\
KG-BERTScore            & \phantom{-}0.485  & \phantom{-}0.428                   & \phantom{-}0.507                   & \phantom{-}0.347                   \\
MS-COMET-QE-22          & \phantom{-}0.483  & \phantom{-}0.488                   & \phantom{-}0.411                   & \phantom{-}0.257                   \\
UniTE-src               & \colorbox[HTML]{B2EAB1}{\textbf{\phantom{-}0.658}}  & \phantom{-}0.445                   & \phantom{-}0.582                   & \phantom{-}0.328                   \\ \midrule
MATESE                  & -0.281 & \phantom{-}n/a                   & n/a & n/a \\
MEE                     & -0.078 & n/a & n/a & n/a \\
MEE2                    & \phantom{-}0.340  & n/a & n/a & n/a \\
MEE4                    & \phantom{-}0.391  & n/a & n/a & n/a \\
REUSE     & \phantom{-}0.430  & n/a & n/a & n/a \\
MATESE-QE & -0.313 & n/a & n/a & n/a \\ 

\bottomrule
\end{tabular}
\caption{Average Kendall’s tau-like correlation results grouped by language pairs: trained language pairs (en-de, en-ru, zh-en), from English (en-x), into English (x-en) and language pairs not involving English (x-y).  The horizontal lines delimit baseline metrics (top), all language pairs participating reference-based metrics (second), all language pairs participating reference-free metrics (third) and trained language pairs only metrics (bottom). The best result for each category is denoted by bold text with a green highlight.}
\label{tab:analysis_langpairs}
\end{table}

Another possible way to evaluate the metrics' performance is not to look at the phenomena but rather at the results on different language pairs. Since \textsc{ACES} covers 146 language pairs and for some of these language pairs we only have very few examples, we decide to split this analysis into four main categories: 

\begin{itemize}
    \item \textbf{trained:} language pairs for which this year's WMT metrics shared task provided training material (en-de, en-ru and zh-en). This category also allows us to analyse the metrics that only cover these specific language pairs and not the full set of language pairs in \textsc{ACES}.
    \item \textbf{en-x:} language pairs where the source language is English.
    \item \textbf{x-en:} language pairs where the target language is English.
    \item \textbf{x-y:} all remaining language pairs, where neither the source language nor the target language are English.
\end{itemize}

Table~\ref{tab:analysis_langpairs} shows the results for all metrics. It is important to note that the results for different language pair categories cannot be directly compared because the examples and covered phenomena categories are not necessarily the same. However, we can compare metrics on each of the language pair groups individually. First, it can again be observed that most submitted metrics outperform the baseline metrics (first group). This shows that the field is advancing and MT evaluation metrics have improved since last year (i.e. 2021).

Interestingly, the six metrics that only scored the trained language pairs (last group in the table) do not outperform the other metrics on the ``trained'' category. Note, however, that the \textsc{MEE*} metrics and \textsc{REUSE} are unsupervised metrics and that the \textsc{MATESE} metrics only used MQM training data. Therefore, we cannot comment on whether creating metrics that are specific to a language pair would result in better metrics. In any case, our findings in Section \ref{subsec:zero-shot} suggest that generalisation to unseen language pairs is generally quite good for the multilingual metrics which might be a more desirable property than increased performance on specific language pairs.

\section{Analysis}
\label{sec:analysis}
Aside from high-level evaluations of which metrics perform best, we are mostly interested in metric-spanning weaknesses that we can identify using \textsc{ACES}. This section shows an analysis of three general questions that we aim to answer using \textsc{ACES}.

\subsection{How sensitive are metrics to the source?}
\label{subsec:source-relevance}
\begin{table*}[ht]
    \centering
    \small
    \resizebox{\linewidth}{!}{%

    \begin{tabular}{lcccccccccc}
    \toprule
    & \multicolumn{2}{c}{\hyperref[subsec:discourse_connectives]{\textbf{since}}} & \multicolumn{2}{c}{\hyperref[subsec:gender_in_occupation_names]{\textbf{female}}}  & \multicolumn{2}{c}{\hyperref[subsec:gender_in_occupation_names]{\textbf{male}}} & \multicolumn{2}{c}{\hyperref[subsec:wsd]{\textbf{wsd}}} &\\
    \cmidrule(lr){2-3} \cmidrule(lr){4-5} \cmidrule(lr){6-7} \cmidrule(lr){8-9} \\
    & \textbf{causal} & \textbf{temp.} & \textbf{anti.} & \textbf{pro.} & \textbf{anti.} & \textbf{pro.} & \textbf{freq.} & \textbf{infreq.} & \textbf{AVG}\\
    \cmidrule(lr){2-2} \cmidrule(lr){3-3} \cmidrule(lr){4-4} \cmidrule(lr){5-5} \cmidrule(lr){6-6}  \cmidrule(lr){7-7} \cmidrule(lr){8-8} \cmidrule(lr){9-9} \cmidrule(lr){10-10} \\
    \textit{\textbf{Examples}} & \textit{106} & \textit{106} & \textit{1000} & \textit{806} & \textit{806} & \textit{1000} & \textit{471} & \textit{471} & \textit{4766}\\
    \midrule
   BERTScore & -0.434 & \phantom{-}0.434 & -0.614 & -0.216 & \phantom{-}0.208 & \phantom{-}0.618 & \phantom{-}0.214 & -0.223 & -0.001\\
   COMET-20 & -0.019 & \phantom{-}0.302 & -0.622 & -0.370 & \colorbox[HTML]{B2EAB1}{\textbf{\phantom{-}0.586}} & \phantom{-}0.772 & \phantom{-}0.202 & -0.079 & \phantom{-}0.097\\
   COMET-22 & -0.415 & \phantom{-}0.792 & \colorbox[HTML]{B2EAB1}{\textbf{\phantom{-}0.940}} & \colorbox[HTML]{B2EAB1}{\textbf{\phantom{-}1.000}} & -0.628 & \phantom{-}0.374 & \colorbox[HTML]{B2EAB1}{\textbf{\phantom{-}0.558}} & \colorbox[HTML]{B2EAB1}{\textbf{\phantom{-}0.040}} & \colorbox[HTML]{B2EAB1}{\textbf{\phantom{-}0.333}}\\
   metricx\_xxl\_DA\_2019 & -0.849 & \phantom{-}0.811 & -0.944 & -0.228 & \phantom{-}0.233 & \colorbox[HTML]{B2EAB1}{\textbf{\phantom{-}0.942}} & \phantom{-}0.032 & -0.028 & -0.004\\
   metricx\_xxl\_MQM\_2020 & -1.000 & \colorbox[HTML]{B2EAB1}{\textbf{\phantom{-}1.000}} & -0.878 & \phantom{-}0.002 & -0.007 & \phantom{-}0.884 & \phantom{-}0.083 & -0.100 & -0.002\\
   MS-COMET-22 & -0.604 & \phantom{-}0.623 & \phantom{-}0.296 & \phantom{-}0.640 & -0.342 & \phantom{-}0.046 & \phantom{-}0.316 & -0.155 & \phantom{-}0.102\\
   UniTE & \colorbox[HTML]{B2EAB1}{\textbf{\phantom{-}0.038}} & -0.075 & -0.890 & -0.213 & \phantom{-}0.377 & \phantom{-}0.934 & \phantom{-}0.270 & -0.223 & \phantom{-}0.027\\
   \midrule
    COMET-QE & -1.000 & \colorbox[HTML]{B2EAB1}{\textbf{\phantom{-}0.981}} & \phantom{-}0.450 & \phantom{-}0.871 & -0.854 & -0.382 & \phantom{-}0.244 & -0.210 & \phantom{-}0.013 \\
   COMET-Kiwi & -0.245 & \phantom{-}0.943 & \phantom{-}0.964 & \phantom{-}0.978 & \phantom{-}0.794 & \colorbox[HTML]{B2EAB1}{\textbf{\phantom{-}0.938}} & \phantom{-}0.648 & \colorbox[HTML]{B2EAB1}{\textbf{\phantom{-}0.363}} & \colorbox[HTML]{B2EAB1}{\textbf{\phantom{-}0.673}} \\
   Cross-QE & \phantom{-}0.208 & \phantom{-}0.830 & \colorbox[HTML]{B2EAB1}{\textbf{\phantom{-}0.976}} & \colorbox[HTML]{B2EAB1}{\textbf{\phantom{-}0.995}} & -0.337 & \phantom{-}0.364 & \colorbox[HTML]{B2EAB1}{\textbf{\phantom{-}0.762}} & \phantom{-}0.355 & \phantom{-}0.519 \\
    HWTSC-Teacher-Sim & -0.453 & \phantom{-}0.717 & \phantom{-}0.916 & \phantom{-}0.772 & -0.283 & -0.360 & \phantom{-}0.295 & \phantom{-}0.079 & \phantom{-}0.210 \\
    KG-BERTScore & \colorbox[HTML]{B2EAB1}{\textbf{\phantom{-}0.453}} & \phantom{-}0.830 & \phantom{-}0.638 & \phantom{-}0.300 & \colorbox[HTML]{B2EAB1}{\textbf{\phantom{-}0.968}} & \phantom{-}0.682 & \phantom{-}0.295 & \phantom{-}0.079 & \phantom{-}0.531 \\
    MS-COMET-QE-22 & -0.283 & \phantom{-}0.792 & -0.194 & \phantom{-}0.320 & \phantom{-}0.246 & \phantom{-}0.694 & \phantom{-}0.465 & \phantom{-}0.002 & \phantom{-}0.255 \\
    UniTE-src & -0.321 & \phantom{-}0.906 & \colorbox[HTML]{B2EAB1}{\textbf{\phantom{-}0.976}} & \phantom{-}0.980 & \phantom{-}0.171 & \phantom{-}0.736 & \phantom{-}0.622 & \phantom{-}0.346 & \phantom{-}0.552 \\
   \bottomrule
    \end{tabular}}
    \caption{Results on the challenge sets where the good translation can only be identified through the source sentence. Upper block: reference-based metrics, lower block: reference-free metrics. Best results for each phenomenon and each group of models is marked in bold and green and the average over all can be seen in the last column.}
    \label{tab:src_disambig}
\end{table*}

We designed our challenge sets for the type of \hyperref[sec:source-disambig]{ambiguous translation} in a way that the correct translation candidate given an ambiguous reference can only be identified through the source sentence. Here, we present a targeted evaluation intended to provide some insights into how important the source is for different metrics. We exclude all metrics that do not take the source as input, all metrics that do not cover all language pairs, and the smaller versions of \textsc{Metric-X} (metricx\_xl\_DA\_2019 and metricx\_xl\_MQM\_2020) from this analysis. This leaves us with 
seven reference-based metrics and seven reference-free metrics. Table~\ref{tab:src_disambig} shows the detailed results of each metric on the considered phenomena.

The most important finding is that the reference-free metrics generally perform much better on these challenge sets than the reference-based metrics. This indicates that reference-based metrics rely too much on the reference. Interestingly, most of the metrics that seem to ignore the source do not randomly guess the correct translation (which is a valid alternative choice when the correct meaning is not identified via the source) but rather they strongly prefer one phenomenon over the other. For example, several metrics show a gender bias either towards female \hyperref[subsec:gender_in_occupation_names]{occupation names} (female correlations are high, male low) or male occupation names (vice versa). Likewise, most metrics prefer translations with frequent senses for the \hyperref[subsec:wsd]{word-sense disambiguation challenge sets}, although the difference between frequent and infrequent is not as pronounced as for gender.

\begin{table}[ht!]
    \centering
    \small
    \begin{tabular}{rc}
\toprule
    & \textbf{corr. gain}\\
\midrule
   BERTScore & 0.002 \\
   COMET-20 & 0.060 \\	
   COMET-22 & \textbf{0.190} \\
   metricx\_xxl\_DA\_2019 & 0.012\\
   metricx\_xxl\_MQM\_2020 & -0.016\\
   MS-COMET-22 & 0.050\\
   UniTE & 0.042\\
\midrule
   COMET-QE & 0.018 \\
   COMET-Kiwi & \textbf{0.338} \\
   Cross-QE & 0.292 \\
   HWTSC-Teacher-Sim & 0.154 \\
   KG-BERTScore & 0.154 \\
   MS-COMET-QE-22 & 0.196 \\
   UniTE-src & 0.216 \\
   \bottomrule
    \end{tabular}
    \caption{Results on the \hyperref[subsec:real-world-commonsense]{real-world knowledge commonsense challenge set} with reference-based metrics in the upper block and reference-free metrics in the lower block. The numbers are computed as the difference between the correlation with the subordinate clause in the source and the correlation without the subordinate clause in the source. Largest gains are bolded.}
    \label{tab:corr_gain}
\end{table}

Only metrics that look at the source and exhibit fewer such preferences can perform well on average on this collection of challenge sets. \textsc{COMET-22} performs best out of the reference-based metrics and \textsc{COMET-Kiwi} performs best of all reference-free metrics. It is noteworthy that there is still a considerable gap between these two models, suggesting that reference-based models should pay more attention to the source when a reference is ambiguous to reach the performance of reference-free metrics.

This finding is also supported by our \hyperref[subsec:real-world-commonsense]{real-world knowledge commonsense challenge set}. If we compare the scores on the examples where the subordinate clauses are missing from both the source and the reference to the ones where they are only missing from the reference, we can directly see the effect of disambiguation through the source. The corresponding correlation gains are shown in Table~\ref{tab:corr_gain}. All reference-based model correlation scores improve less than most reference-free correlations when access to the subordinate clause is given through the source. This highlights again that reference-based metrics do not give enough weight to the source sentence.

\subsection{How much do metrics rely on surface-overlap with the reference?}
\label{subsec:surface-relevance}
\begin{figure}
    \centering
    \includegraphics[width=0.48\textwidth]{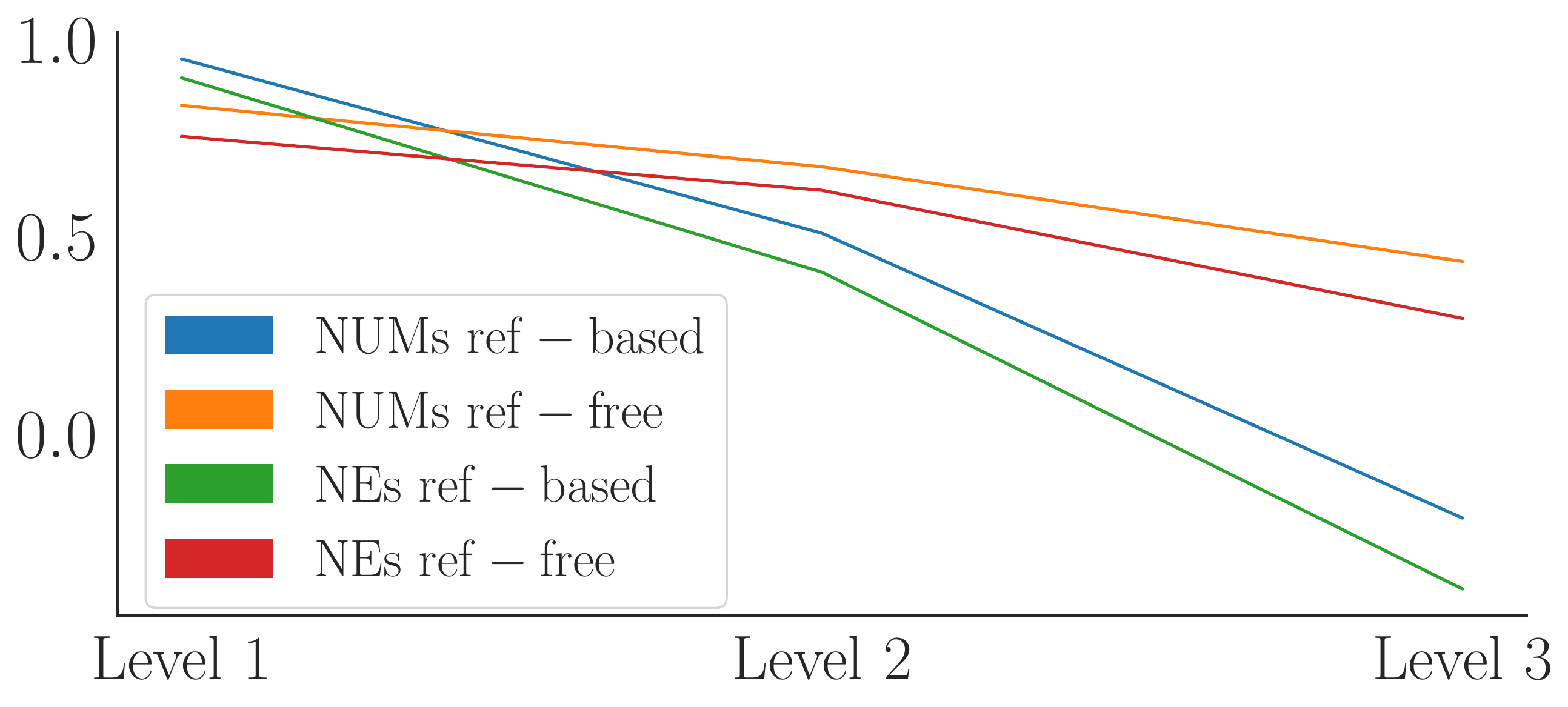}
    \caption{Decrease in correlation for reference-based and reference-free metrics on the \hyperref[subsec:levels]{named entity and number hallucination challenge sets}.}
    \label{fig:corr_decrease}
\end{figure}

Another question we are interested in is whether neural reference-based metrics still rely on surface-level overlap with the reference. For this analysis, we use the dataset we created for \hyperref[subsec:levels]{hallucinated named entities and numbers}. We take the average correlation for all reference-based metrics\footnote{Excluding surface-level baseline metrics: \textsc{BLEU},  \textsc{spBLEU} and \textsc{chrf}.} and the average correlation of all reference-free metrics that cover all languages and plot the decrease in correlation with increasing surface-level similarity of the incorrect translation to the reference. The result can be seen in Figure~\ref{fig:corr_decrease}.

We can see that on average reference-based metrics have a much steeper decrease in correlation than the reference-free metrics as the two translation candidates become more and more lexically diverse and the surface overlap between the incorrect translation and the reference increases. This indicates a possible weakness of reference-based metrics: If one translation is lexically similar to the reference but contains a grave error while others are correct but share less surface-level overlap with the reference, the incorrect translation may still be preferred.

We also show that this is the case for the challenge set where we use an adversarial paraphrase from PAWS-X that shares a high degree of \hyperref[subsec:lexical-overlap]{lexical overlap} with the reference but does not have the same meaning as an incorrect translation. On average, the reference-based metrics only reach a correlation of 0.05 ± 0.12 on this challenge set, whereas the reference-free metrics reach a correlation of 0.23 ± 0.15. This shows that reference-based metrics are less robust when the incorrect translation has high lexical overlap with the reference.

Finally, we can also see a clear effect of surface-level overlap with the source on three real error challenge sets where we have different versions of the good translation: some where the error was corrected with the corresponding correct token from the reference and some where the error was corrected with a synonym for the correct token from the reference. As seen in Table \ref{tab:my_label}, the reference-based metrics show a much larger difference in correlation between the challenge sets with reference- copied good translations and the challenge sets with the synonymous good translations, than the reference-free metrics. For example, for the \hyperref[subsec:real_hallucination]{hallucination test set}, reference-free metrics have very similar average performance when the good translation contains the same word as the reference vs. when it contains a synonym ($\delta$ of +0.04). On the other hand, the reference-based metrics lose on average -0.22 in correlation when the good translation contains the synonym rather than the same word as the reference. Based on all of these results, we conclude that even though state-of-the-art reference-based MT evaluation metrics are not only reliant on surface-level overlap anymore, such overlap still considerably influences their predictions.

\begin{table}[]
    \centering
    \small
    \begin{tabular}{ccc}
    \toprule
         & reference-based & reference-free \\
        \midrule
        \hyperref[subsec:real_hallucination]{hallucination} & -0.22 ± 0.16 & +0.04 ± 0.07 \\
        \hyperref[subsec:real_overly_literal]{overly-literal} & -0.32 ± 0.16
        & +0.12 ± 0.09\\
        \hyperref[subsec:real_untranslated]{untranslated} & -0.44 ± 0.18 & +0.03 ± 0.06\\
    \bottomrule
    \end{tabular}
    \caption{Average correlation difference and standard deviation between the challenge sets with reference-copied good translations and the challenge sets with the synonymous good translations.}
    \label{tab:my_label}
\end{table}

\subsection{Do multilingual embeddings help design better metrics?}
\label{subsec:mutlilingual embeddings}
As the community moves towards building metrics that use multilingual encoders, we investigate if some (un)desirable properties of multilingual embeddings are propagated in these metrics. 

\subsubsection{Zero-shot Performance}
\label{subsec:zero-shot}

Similar to \citet{kocmi-etal-2021-ship}, we investigate whether there is a difference in the performance of metrics on our challenge sets when evaluated on non-WMT language pairs \textit{i.e.} language pairs unseen during the training of the metrics. For this analysis, we include only those metrics for which the training data consisted of some combination of WMT human evaluation data. As different metrics used data from different years, we consider an intersection of languages across these years as WMT language pairs. For a fair comparison, we consider a subset of examples from those phenomena where we have least 100 examples in WMT languages and 100 examples in non-WMT languages, irrespective of the number of examples per individual language pair. We report some of the phenomena in Table~\ref{tab:zero-shot}, where metrics are compared in terms of the correlation difference between the performance on WMT and non-WMT language pairs (see Appendix~\ref{app:zero-shot} for the original WMT and non-WMT correlation scores and the list of language pairs).
\begin{table}[h]
\centering
\small
    \resizebox{\linewidth}{!}{%

\begin{tabular}{lccc}
\toprule
 &
  \multicolumn{1}{l}{\begin{tabular}[c]{@{}l@{}}\hyperref[sec:antonym]{antonym-}\\ \hyperref[sec:antonym]{replacement}\end{tabular}} &
  \multicolumn{1}{l}{\begin{tabular}[c]{@{}l@{}}\hyperref[subsec:real-world-commonsense]{real-world}\\ \hyperref[subsec:real-world-commonsense]{knowledge}\\ \hyperref[subsec:real-world-commonsense]{commonsense}\end{tabular}} &
  \multicolumn{1}{l}{\hyperref[sec:nonsense]{nonsense}} \\ \midrule
\textit{Examples}           & \textit{131}    & \textit{201}    & \textit{239}    \\ \midrule
BERTScore               & \phantom{-} 0.032  & -0.054 & \phantom{-} 1.469  \\
BLEURT-20               & \phantom{-} 0.032  & \phantom{-} 0.201  & \phantom{-} 0.350  \\
COMET-20                & \phantom{-} 0.048  & \phantom{-} 0.067  & \phantom{-} 1.021  \\
COMET-QE                & -0.048 & -0.188 & -0.294 \\
COMET-22                & \phantom{-} 0.080  & \phantom{-} 0.027  & \phantom{-} 0.531  \\ \midrule
metricx\_xl\_DA\_2019   & -0.032 & -0.054 & \phantom{-} 0.434  \\ 
metricx\_xl\_MQM\_2020  & -0.048 & -0.094 & \phantom{-} 0.182  \\
metricx\_xxl\_DA\_2019  & \phantom{-} 0.016  & -0.040 & \phantom{-} 0.266  \\
metricx\_xxl\_MQM\_2020 & \phantom{-} 0.064  & -0.067 & \phantom{-} 0.196  \\
UniTE-ref               & -0.032 & \phantom{-} 0.013  & \phantom{-} 0.238  \\
UniTE                   & \phantom{-} 0.080  & \phantom{-} 0.000  & \phantom{-} 0.643  \\ \midrule
COMETKiwi               & \phantom{-} 0.048  & -0.027 & \phantom{-} 0.042  \\
Cross-QE                & \phantom{-} 0.064  & \phantom{-} 0.188  & \phantom{-} 0.182  \\
HWTSC-Teacher-Sim       & \phantom{-} 0.208  & \phantom{-} 0.081  & \phantom{-} 0.350  \\
UniTE-src               & \phantom{-} 0.096  & \phantom{-} 0.161  & -0.028 \\ \bottomrule
\end{tabular}}
\caption{Correlation difference between the performance of WMT and non-WMT language pairs reported for trained metrics across a subset of examples. $\delta$= $\tau_{\text{WMT}}$ - $\tau_{\text{non WMT}}$. WMT language pairs consist of a subset of languages seen during training of the metrics, while non-WMT language pairs are unseen. Results show that the metrics are able to generalise to unseen languages. }
\label{tab:zero-shot}
\end{table}

We draw similar conclusions to \citet{kocmi-etal-2021-ship}, namely that trained metrics are not over-fitted to the WMT language pairs. We observe that the median difference of $\tau$ between WMT and non-WMT language pairs is 0.056, indicating a good generalisation to unseen languages. We still note that performance on the phenomena is variable when we compare the results on WMT language pairs versus non-WMT language pairs. In the case of \hyperref[subsec:real-world-commonsense]{real-world knowledge commonsense}, performance is slightly better on the non-WMT language pairs\footnote{We also observe better performance on non-WMT language pairs for the \hyperref[sec:wrong_language]{similar language high phenomenon}.}, while the opposite is (generally) true for the \hyperref[sec:antonym]{antonym replacement} and, especially, the \hyperref[sec:nonsense]{nonsense} phenomena for certain metrics. Further analysis is required to better understand metric behaviour on zero-shot language pairs, especially considering that some of the analysed non-WMT language pairs have a target language that is also the target language in at least one of the WMT language pairs (e.g. English).

\subsubsection{Language Dependent Representations}
\begin{figure}
    \centering
    \includegraphics[width=0.46\textwidth]{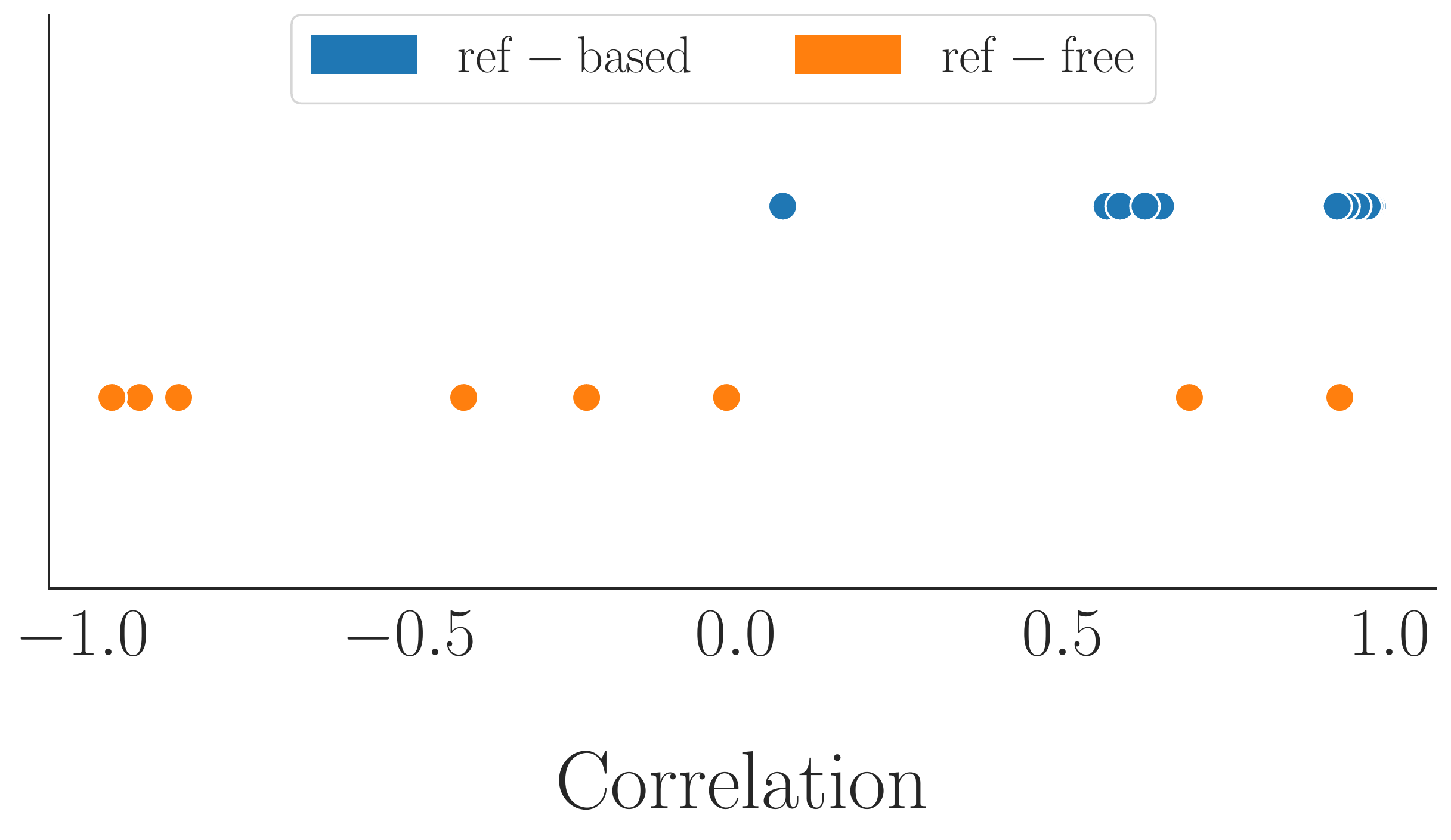}
    \caption{Correlation of reference-based metrics (blue) and reference-free metrics (orange) on the  \hyperref[subsec:sent-untranslated]{sentence-level untranslated test challenge set}.}
   \label{fig:corr_copy_src}
\end{figure}

Multilingual models often learn cross-lingual representations by abstracting away from language-specific information \citep{wu-dredze-2019-beto}. We are interested in whether the representations are still language-dependent in neural MT evaluation metrics which are trained on such models. For this analysis, we look at the \hyperref[subsec:sent-untranslated]{sentence-level untranslated text} challenge set (see Figure \ref{fig:corr_copy_src}) and \hyperref[sec:wrong_language]{wrong language phenomena} (see Table~\ref{tab:analysis_overview}).
We only consider metrics that provided scores for examples in all language pairs.

Figure \ref{fig:corr_copy_src} shows the correlations for all reference-based and reference-free metrics. Unsurprisingly, some 
reference-free metrics struggle considerably on this challenge set and almost always prefer the copied source to the real translation. The representations of the source and the incorrect translation are identical, leading to a higher surface and embedding similarity, and thus a higher score. We do, however, find some exceptions to this trend - \textsc{COMET-Kiwi} and \textsc{MS-COMET-QE-22} both have a high correlation on  \hyperref[subsec:sent-untranslated]{sentence-level untranslated text}. This suggests that these metrics could have learnt language-dependent representations. 

Most reference-based metrics have good to almost perfect correlation and can identify the copied source quite easily. As reference-based metrics tend to ignore the source (see Section~\ref{subsec:surface-relevance}), the scores are based on the similarity between the reference and the MT output. In this challenge set, the similarity between the good-translation and the reference is likely to be higher than the incorrect-translation and the reference. The former MT output is in the same language as the reference and will have more surface level overlap. We believe the reference here acts as grounding.

However, this grounding property of the reference is only robust when the source and reference languages are dissimilar, as is the case with language pairs in the \hyperref[subsec:sent-untranslated]{sentence-level untranslated text} challenge set. We find that reference-based metrics struggle on \hyperref[sec:wrong_language]{wrong language phenomena}  (see Table \ref{tab:analysis_overview}) where the setup is similar, but now the incorrect translation and the reference are from similar languages (e.g. one is in Hindi and the other is in Marathi). Naturally, there will be surface level overlap between the reference and both the good-translation and the incorrect-translation. For example, both Marathi and Hindi use named entities with identical surface form, and so these will appear in the reference and also in both the good-translation and the incorrect-translation. Thus, the semantic content drives the similarity scores between the MT outputs and the references. It is possible that the human translation in the similar language (labelled as the incorrect-translation) has a closer representation to the human reference because in the MT output (labelled as the good-translation) some semantic information may be lost. We leave further investigation of this for future work.

While multilingual embeddings help in effective zero-shot transfer to new languages, some properties of the multilingual representation space may need to be altered to suit the task of machine translation evaluation. 

\section{Recommendations}
\label{sec:recommendations}
Based on the metrics results on \textsc{ACES} and our analysis, we derived the following list of recommendations for future MT evaluation metric development: 

\textbf{No metric to rule them all:} Both the evaluation on phenomena and on language pair categories in Section~\ref{sec: Results} showed that there is no single best-performing metric. This divergence is likely to become even larger if we evaluate metrics on different domains. For future work on MT evaluation, it may be worthwhile thinking about how different metrics can be combined to make robust decisions as to which is the best translation. This year's submissions to the metrics shared task already suggest that work in that direction is ongoing as some groups submitted metrics that combined ensembles of models or multiple components (\textsc{COMET-22}, \textsc{COMET-Kiwi}, \textsc{KG-BERTScore}, \textsc{MEE*}, \textsc{REUSE}).

\textbf{The source matters:} Our analysis in Section~\ref{subsec:source-relevance} highlighted that many reference-based metrics that take the source as input do not consider it enough. Cases where the correct translation can only be identified through the source are currently better handled by reference-free metrics. This is a serious shortcoming of reference-based metrics and should be addressed in future research, also considering that many reference-based metrics do not even take the source as input.

\textbf{Surface-overlap still prevails:} In Section~\ref{subsec:surface-relevance}, we showed that despite moving beyond only surface-level comparison to the reference, most reference-based metric scores are still considerably influenced by surface-level overlap. We expect future metrics to use more lexically diverse references in their training regime to mitigate this issue.

\textbf{Multilingual embeddings are not perfect:} Some properties of multilingual representations, especially, being language-agnostic, can result in undesirable effects on MT evaluation (Section ~\ref{subsec:mutlilingual embeddings}). It could be helpful for future metrics to incorporate strategies to explicitly model additional language-specific information.

\section{Conclusion}
We presented \textsc{ACES}, a translation accuracy challenge set based on the MQM ontology. \textsc{ACES} consists of 36,476 examples covering 146 language pairs and representing challenges
from 68 phenomena. We used \textsc{ACES} to evaluate the baseline and submitted metrics from the WMT 2022 metrics shared task. Our overview of metric performance at the phenomena and language levels in Section~\ref{sec: Results} reveals that there is no single best-performing metric. The more fine-grained analyses in Section~\ref{sec:analysis} highlight that 1) many reference-based metrics that take the source as input do not consider it enough, 2) most
reference-based metric scores are still considerably
influenced by surface overlap with the reference, and 3) the use of multilingual embeddings can have undesirable effects on MT evaluation.

We recommend that these shortcomings of existing metrics
be addressed in future research, and that metric developers should consider a) combining metrics with different strengths, e.g. in the form of ensemble models, b) developing metrics that give more weight to the source and less to surface-level overlap with the reference, and c) incorporating strategies to explicitly model additional language-specific information (rather than simply relying on multilingual embeddings). 

We have made \textsc{ACES} publicly available and hope that it will provide a useful benchmark for MT evaluation metric developers in the future.

\section*{Limitations}
The \textsc{ACES} challenge set exhibits a number of biases. Firstly, there is greater coverage in terms of phenomena and number of examples for the en-de and en-fr language pairs. This is in part due to the manual effort required to construct examples for some phenomena, in particular those belonging to the \hyperref[sec:discourse]{discourse-level} and \hyperref[sec:real-world-knowledge]{real-world knowledge} categories. Further, our choice of language pairs is also limited to the ones available in XLM-R. Secondly, \textsc{ACES} contains more examples for those phenomena for which examples could be generated automatically, compared to those that required manual construction/filtering. Thirdly, some of the automatically generated examples require external libraries which are only available for a few languages (e.g. Multilingual Wordnet). Fourthly, the focus of the challenge set is on accuracy errors. We leave the development of challenge sets for fluency errors to future work.

As a result of using existing datasets as the basis for many of the examples, errors present in these datasets may be propagated through into ACES. Whilst we acknowledge that this is undesirable, in our methods for constructing the \textit{incorrect translation} we aim to ensure that the quality of the \textit{incorrect translation} is always worse than the corresponding \textit{good translation}.

The results and analyses presented in the paper exclude those metrics submitted to the WMT 2022 metrics shared task that provide only system-level outputs. We focus on metrics that provide segment-level outputs as this enables us to provide a broad overview of metric performance on different phenomenon categories and to conduct fine-grained analyses of performance on individual phenomena. For some of the fine-grained analyses, we apply additional constraints based on the language pairs covered by the metrics, or whether the metrics take the source as input, to address specific questions of interest. As a result of applying some of these additional constraints, our investigations tend to focus more on high and medium-resource languages than on low-resource languages. We hope to address this shortcoming in future work.

\section*{Ethics Statement}
Some examples within the challenge set exhibit biases, however this is necessary in order to expose the limitations of existing metrics. Wherever external help was required in verifying translations, the annotators were compensated at a rate of £15/hour. 
Our challenge set is based on publicly available datasets and will be released for future use. 

\section*{Acknowledgements}
We thank the organisers of the WMT 2022 Metrics task for setting up this shared task and for their feedback throughout the process, and the shared task participants for scoring our challenge sets with their systems. We are grateful to Stephanie Droop, Octave Mariotti, and Kenya Murakami for helping us with the annotations. We thank the StatMT group at Edinburgh, especially Alexandra Birch, Barry Haddow, and Ulrich Germann, and the attendees at the MT Marathon 2022 for their valuable feedback. We thank Janis Goldzycher, Mark Steedman, Rico Sennrich, and the anonymous reviewers for their insightful comments and suggestions. This work was supported in part by the UKRI Centre for Doctoral Training in Natural Language Processing, funded by the UKRI (grant EP/S022481/1) and the University of Edinburgh (Moghe), by the Swiss National Science Foundation (project MUTAMUR; no. 176727) (Amrhein) and by the ERC H2020 Advanced Fellowship GA 742137 SEMANTAX (Guillou). We also thank Huawei for their support (Moghe).

% Entries for the entire Anthology, followed by custom entries
\bibliography{anthology,custom}
\bibliographystyle{acl_natbib}

\newpage

\appendix

\section{Appendix}
\label{sec:appendix}
 \subsection{Language Codes}

\begin{table}[ht]
\begin{tabular}{ll|ll}
\hline
Code & Language                                                 & Code & Language   \\ \hline
af   &  Afrikaans & ja   & Japanese   \\
ar   &  Arabic    & ko   & Korean     \\
be   & Belarusian                                               & lt   & Lithuanian    \\
bg   & Bulgarian                                                & lv   & Latvian    \\
ca   & Catalan                                                  & mr   & Marathi    \\
cs   & Czech                                                    & nl   & Dutch      \\
da   & Danish                                                   & no   & Norwegian  \\
de   & German                                                   & pl   & Polish     \\
el   & Greek                                                    & pt   & Portuguese \\
en   & English                                                  & ro   & Romanian   \\
es   & Spanish                                                  & ru   & Russian    \\
et   & Estonian                                                 & sk   & Slovak     \\
fa   & Persian                                                  & sl   & Slovenian  \\
fi   & Finnish                                                  & sr   & Serbian    \\
fr   & French                                                   & sv   & Swedish    \\
ga   & Irish                                                    & sw   & Swahili    \\
gl   & Galician                                                 & ta   & Tamil      \\
he   & Hebrew                                                   & th   & Thai       \\
hi   & Hindi                                                    & tr   & Turkish    \\
hr   & Croatian                                                 & uk   & Ukranian   \\
hu   & Hungarian                                                & ur   & Urdu       \\
hy   & Armenian                                                 & vi   & Vietnamese \\
id   & Indonesian                                               & wo   & Wolof      \\
it   & Italian                                                  & zh   & Chinese    \\ \hline      
\end{tabular}
\caption{ISO 2-Letter language codes of the languages included in the challenge set}
\label{lang_codes}
\end{table}

\subsection{Permitted Unit Conversions}
\label{app:allowed-units}

We allow the following \hyperref[subsec:units]{unit conversions} for the challenge set that covers such errors:\\

\noindent \textbf{Distance}:
\begin{itemize}
    \setlength\itemsep{0.05em}
    \item miles $\rightarrow$ metres
    \item kilometres $\rightarrow$ miles
    \item kilometres $\rightarrow$ metres
    \item metres $\rightarrow$ feet
    \item metres $\rightarrow$ yards
    \item feet $\rightarrow$ metres
    \item feet $\rightarrow$ yards
    \item centimetres $\rightarrow$ inches
    \item centimetres $\rightarrow$ millimetres
    \item inches $\rightarrow$ centimetres
    \item inches $\rightarrow$ millimetres
    \item millimetres $\rightarrow$ centimetres
    \item millimetres $\rightarrow$ inches
    \item millimetres $\rightarrow$ inches
\end{itemize}

\noindent \textbf{Speed}:
\begin{itemize}
    \setlength\itemsep{0.05em}
    \item miles per hour $\rightarrow$ kilometres per hour
    \item kilometres per hour $\rightarrow$ miles per hour
    \item kilometres per second $\rightarrow$ miles per second
    \item miles per second $\rightarrow$ kilometres per second
\end{itemize}

\noindent \textbf{Time}:
\begin{itemize}
    \setlength\itemsep{0.05em}
    \item hours $\rightarrow$ minutes
    \item minutes $\rightarrow$ seconds
    \item seconds $\rightarrow$ minutes
    \item days $\rightarrow$ hours
    \item months $\rightarrow$ weeks
    \item weeks $\rightarrow$ days
\end{itemize}

\noindent \textbf{Volume}:
\begin{itemize}
    \setlength\itemsep{0.05em}
    \item barrels $\rightarrow$ gallons
    \item barrels $\rightarrow$ litres
    \item gallons $\rightarrow$ barrels
    \item gallons $\rightarrow$ litres
\end{itemize}

\noindent \textbf{Weight}:
\begin{itemize}
    \setlength\itemsep{0.05em}
    \item kilograms $\rightarrow$ grams
    \item kilograms $\rightarrow$ pounds
    \item grams $\rightarrow$ ounces
    \item ounces $\rightarrow$ grams
\end{itemize}

\noindent \textbf{Area}:
\begin{itemize}
    \item square kilometres $\rightarrow$ square miles
\end{itemize}

\vspace{1em} % Move the following section into the next column

\subsection{Zero Shot Performance Scores}
\label{app:zero-shot}
\begin{table*}[t]
\small
\centering
\begin{tabular}{lcccccc}
\toprule
 &
  \multicolumn{2}{c}{\hyperref[sec:antonym]{\textbf{antonym-replacement}}} &
  \multicolumn{2}{c}{\begin{tabular}[c]{@{}l@{}}\hyperref[subsec:real-world-commonsense]{\textbf{real-world knowledge}}\\\hyperref[subsec:real-world-commonsense]{\textbf{-commonsense}}\end{tabular}} &
  \multicolumn{2}{c}{\hyperref[sec:nonsense]{\textbf{nonsense}}} \\
\midrule
 &
  \multicolumn{1}{c}{WMT} &
  \multicolumn{1}{c}{\begin{tabular}[c]{@{}l@{}}Non-\\  WMT\end{tabular}} &
  \multicolumn{1}{c}{WMT} &
  \multicolumn{1}{c}{\begin{tabular}[c]{@{}l@{}}Non-\\ WMT\end{tabular}} &
  \multicolumn{1}{c}{WMT} &
  \multicolumn{1}{c}{\begin{tabular}[c]{@{}l@{}}Non-\\ WMT\end{tabular}} \\
\midrule
BERTScore               & -0.376 & -0.408 & \phantom{-} 0.007 & \phantom{-} 0.060 & \phantom{-} 0.790  & -0.678 \\
BLEURT-20               & \phantom{-} 0.024  & -0.008 & \phantom{-} 0.396 & \phantom{-} 0.195 & -0.273 & -0.622 \\
COMET-20                & \phantom{-} 0.152  & \phantom{-} 0.104  & \phantom{-} 0.087 & \phantom{-} 0.020 & \phantom{-} 0.706  & -0.315 \\ 
COMET-QE                & \phantom{-} 0.616  & \phantom{-} 0.664  & \phantom{-} 0.168 & \phantom{-} 0.356 & \phantom{-} 0.245  & \phantom{-} 0.538  \\ \midrule
COMET-22                & \phantom{-} 0.744  & \phantom{-} 0.664  & \phantom{-} 0.584 & \phantom{-} 0.557 & \phantom{-} 0.706  & \phantom{-} 0.175  \\
metricx\_xl\_DA\_2019   & \phantom{-} 0.728  & \phantom{-} 0.760  & \phantom{-} 0.570 & \phantom{-} 0.624 & \phantom{-} 0.790  & \phantom{-} 0.357  \\
metricx\_xl\_MQM\_2020  & \phantom{-} 0.888  & \phantom{-} 0.936  & \phantom{-} 0.517 & \phantom{-} 0.611 & \phantom{-} 0.944  & \phantom{-} 0.762  \\
metricx\_xxl\_DA\_2019  & \phantom{-} 0.312  & \phantom{-} 0.296  & \phantom{-} 0.718 & \phantom{-} 0.758 & \phantom{-} 0.706  & \phantom{-} 0.441  \\
metricx\_xxl\_MQM\_2020 & \phantom{-} 0.696  & \phantom{-} 0.632  & \phantom{-} 0.691 & \phantom{-} 0.758 & \phantom{-} 0.930  & \phantom{-} 0.734  \\
UniTE-ref               & \phantom{-} 0.664  & \phantom{-} 0.696  & \phantom{-} 0.409 & \phantom{-} 0.396 & \phantom{-} 0.091  & -0.147 \\
UniTE                   & \phantom{-} 0.632  & \phantom{-} 0.552  & \phantom{-} 0.409 & \phantom{-} 0.409 & \phantom{-} 0.441  & -0.203 \\ \midrule
COMETKiwi               & \phantom{-} 0.744  & \phantom{-} 0.696  & \phantom{-} 0.745 & \phantom{-} 0.772 & \phantom{-} 0.510  & \phantom{-} 0.469  \\
Cross-QE                & \phantom{-} 0.680  & \phantom{-} 0.616  & \phantom{-} 0.638 & \phantom{-} 0.450 & \phantom{-} 0.720  & \phantom{-} 0.538  \\
HWTSC-Teacher-Sim       & \phantom{-} 0.504  & \phantom{-} 0.296  & \phantom{-} 0.248 & \phantom{-} 0.168 & \phantom{-} 0.930  & \phantom{-} 0.580  \\
UniTE-src               & \phantom{-} 0.776  & \phantom{-} 0.680  & \phantom{-} 0.651 & \phantom{-} 0.490 & \phantom{-} 0.524  & \phantom{-} 0.552  \\ 
 \bottomrule
\end{tabular}
\caption{Zero-shot performance of neural metrics on three phenomena to measure the ability of metrics to generalise to new language pairs. WMT language pairs consist of a subset of languages seen during training of the metrics, while non-WMT language pairs are unseen. Results show that the metrics are able to generalise to unseen languages. }
\label{tab:zero-shot-appendix}
\end{table*}
Table~\ref{tab:zero-shot-appendix} contains the Kendall tau-like correlation scores for neural metrics on WMT language pairs (a subset of those seen during training) and non-WMT language pairs (unseen), for three phenomena: \hyperref[sec:antonym]{antonym replacement}, \hyperref[subsec:real-world-commonsense]{real-world knowledge commonsense}, and \hyperref[sec:nonsense]{nonsense}. The table contains the complete set of scores, and complements Table~\ref{tab:zero-shot}, which reports only the difference between the non-WMT and WMT correlation scores. See Section~\ref{subsec:zero-shot} on zero-shot performance. We shall now list the language pairs across the different phenomena: 

\begin{itemize}
    \item[] \textit{Antonym Replacement}\\
    WMT: \hfill de-en\\
    non-WMT: \hfill ko-en, es-en
    \item[] \textit{Real-world Knowledge - Commonsense}\\
    WMT: \hfill de-en, ru-en, en-ru, en-de\\
    non-WMT:\hfill ru-de, fr-ru, ru-fr, de-ru
    \item[] \textit{Nonsense}\\
    WMT:\hfill de-en\\
    non-WMT:\hfill fr-ja, ko-ja, en-ko, ko-en
\end{itemize}

\noindent Note that the subset of examples used in this analysis only consists of mid/high resource language pairs; investigation into the performance on low-resource languages is left for future work.

\subsection{Distribution of Examples Across Language Pairs}
\label{app:language_pair_matrix}
\begin{sidewaystable*}[]
\tiny
\setlength\tabcolsep{3.15 pt}
\renewcommand{\arraystretch}{1.15}
\begin{tabular}{ll|c|c|c|c|c|c|c|c|c|c|c|c|c|c|c|c|c|c|c|c|c|c|c|c|c|c|c|c|c|c|c|c|c|c|c|c|c|c|c|c|c|c|c|c|c|c|c|c|}
\multicolumn{1}{c}{} & \multicolumn{1}{c}{} & \multicolumn{48}{l}{\textbf{tgt $\rightarrow$}} \\

\multicolumn{1}{c}{} & \multicolumn{1}{c}{} & \multicolumn{1}{c}{af} & \multicolumn{1}{c}{ar} & \multicolumn{1}{c}{be} & \multicolumn{1}{c}{bg} & \multicolumn{1}{c}{ca} & \multicolumn{1}{c}{cs} & \multicolumn{1}{c}{da} & \multicolumn{1}{c}{de} & \multicolumn{1}{c}{el} & \multicolumn{1}{c}{en} & \multicolumn{1}{c}{es} & \multicolumn{1}{c}{et} & \multicolumn{1}{c}{fa} & \multicolumn{1}{c}{fi} & \multicolumn{1}{c}{fr} & \multicolumn{1}{c}{ga} & \multicolumn{1}{c}{gl} & \multicolumn{1}{c}{he} & \multicolumn{1}{c}{hi} & \multicolumn{1}{c}{hr} & \multicolumn{1}{c}{hu} & \multicolumn{1}{c}{hy} & \multicolumn{1}{c}{id} & \multicolumn{1}{c}{it} & \multicolumn{1}{c}{ja} & \multicolumn{1}{c}{ko} & \multicolumn{1}{c}{lt} & \multicolumn{1}{c}{lv} & \multicolumn{1}{c}{mr} & \multicolumn{1}{c}{nl} & \multicolumn{1}{c}{no} & \multicolumn{1}{c}{pl} & \multicolumn{1}{c}{pt} & \multicolumn{1}{c}{ro} & \multicolumn{1}{c}{ru} & \multicolumn{1}{c}{sk} & \multicolumn{1}{c}{sl} & \multicolumn{1}{c}{sr} & \multicolumn{1}{c}{sv} & \multicolumn{1}{c}{sw} & \multicolumn{1}{c}{ta} & \multicolumn{1}{c}{th} & \multicolumn{1}{c}{tr} & \multicolumn{1}{c}{uk} & \multicolumn{1}{c}{ur} & \multicolumn{1}{c}{vi} & \multicolumn{1}{c}{wo} & \multicolumn{1}{c}{zh}\\
\hhline{~~------------------------------------------------}
\textbf{src} & af &  &  &  &  &  &  &  &  &  & \cellcolor{red!10}{96} &  &  & \cellcolor{red!10}{25} &  &  &  &  &  &  &  &  &  &  &  &  &  &  &  &  &  &  &  &  &  &  &  &  &  &  &  &  &  &  &  &  &  &  & \\
\hhline{~~------------------------------------------------}
\textbf{$\downarrow$} & ar &  &  &  &  &  &  &  &  &  & \cellcolor{red!25}{361} &  &  &  &  & \cellcolor{red!25}{102} &  &  &  & \cellcolor{red!10}{17} &  &  &  &  &  &  &  &  &  &  &  &  &  &  &  &  &  &  &  &  &  &  &  &  &  &  &  &  & \\
\hhline{~~------------------------------------------------}
 & be &  &  &  &  &  &  &  &  &  & \cellcolor{red!10}{67} &  &  &  &  &  &  &  &  &  &  &  &  &  &  &  &  &  &  &  &  &  &  &  &  &  &  &  &  &  &  &  &  &  &  &  &  &  & \\
\hhline{~~------------------------------------------------}
 & bg &  &  &  &  &  &  &  &  &  & \cellcolor{red!25}{393} &  &  &  &  &  &  &  &  &  &  &  &  &  &  &  &  & \cellcolor{red!10}{40} &  &  &  &  &  &  &  &  &  &  &  &  &  &  &  &  &  &  &  &  & \\
\hhline{~~------------------------------------------------}
 & ca &  &  &  &  &  &  &  &  &  & \cellcolor{red!10}{79} & \cellcolor{red!25}{175} &  &  &  &  &  &  &  &  &  &  &  &  &  &  &  &  &  &  &  &  &  &  &  &  &  &  &  &  &  &  &  &  &  &  &  &  & \\
\hhline{~~------------------------------------------------}
 & cs &  &  &  &  &  &  &  &  &  & \cellcolor{red!10}{85} &  &  &  &  &  &  &  &  &  &  &  &  &  &  &  &  &  &  &  &  &  &  &  &  &  &  &  &  &  &  &  &  &  &  &  &  &  & \\
\hhline{~~------------------------------------------------}
 & da &  &  &  &  &  &  &  &  &  & \cellcolor{red!10}{83} &  &  &  &  &  &  &  &  &  &  &  &  &  &  &  &  &  &  &  &  &  &  &  &  &  &  &  &  &  &  &  &  &  &  &  &  &  & \\
\hhline{~~------------------------------------------------}
 & de &  &  &  &  &  &  &  &  &  & \cellcolor{red!40}{4163} & \cellcolor{red!10}{84} &  &  &  & \cellcolor{red!25}{394} &  &  &  &  &  &  &  &  &  & \cellcolor{red!25}{113} & \cellcolor{red!10}{63} &  &  &  &  &  &  &  &  & \cellcolor{red!25}{104} &  &  &  &  &  &  &  &  &  &  &  &  & \cellcolor{red!10}{75}\\
\hhline{~~------------------------------------------------}
 & el &  &  &  &  &  &  &  &  &  & \cellcolor{red!25}{387} &  &  &  &  &  &  &  &  &  &  &  &  &  &  &  &  &  &  &  &  &  &  &  &  &  &  &  &  &  &  &  &  &  &  &  &  &  & \\
\hhline{~~------------------------------------------------}
 & en & \cellcolor{red!10}{25} & \cellcolor{red!10}{5} & \cellcolor{red!10}{6} & \cellcolor{red!10}{15} & \cellcolor{red!25}{347} & \cellcolor{red!25}{368} & \cellcolor{red!10}{46} & \cellcolor{red!40}{6964} & \cellcolor{red!10}{21} &  & \cellcolor{red!25}{725} & \cellcolor{red!10}{25} & \cellcolor{red!10}{20} & \cellcolor{red!10}{12} & \cellcolor{red!25}{800} &  & \cellcolor{red!10}{16} & \cellcolor{red!10}{18} & \cellcolor{red!25}{343} & \cellcolor{red!10}{27} & \cellcolor{red!10}{44} & \cellcolor{red!10}{3} & \cellcolor{red!10}{31} & \cellcolor{red!10}{10} & \cellcolor{red!25}{430} & \cellcolor{red!25}{545} & \cellcolor{red!10}{17} & \cellcolor{red!10}{19} & \cellcolor{red!10}{52} & \cellcolor{red!10}{50} & \cellcolor{red!10}{53} & \cellcolor{red!25}{349} & \cellcolor{red!10}{44} & \cellcolor{red!10}{46} & \cellcolor{red!25}{698} & \cellcolor{red!10}{27} & \cellcolor{red!10}{45} & \cellcolor{red!10}{15} & \cellcolor{red!10}{39} &  & \cellcolor{red!10}{1} &  & \cellcolor{red!10}{10} & \cellcolor{red!10}{16} & \cellcolor{red!10}{10} & \cellcolor{red!10}{25} &  & \cellcolor{red!25}{333}\\
\hhline{~~------------------------------------------------}
 & es &  &  &  &  & \cellcolor{red!10}{88} &  &  & \cellcolor{red!10}{64} &  & \cellcolor{red!40}{1263} &  &  &  &  & \cellcolor{red!25}{125} &  &  &  &  &  &  &  &  &  & \cellcolor{red!25}{117} & \cellcolor{red!10}{74} &  &  &  &  &  &  &  &  &  &  &  &  &  &  &  &  &  &  &  &  &  & \cellcolor{red!10}{67}\\
\hhline{~~------------------------------------------------}
 & et &  &  &  &  &  &  &  &  &  & \cellcolor{red!10}{70} &  &  &  &  &  &  &  &  &  &  &  &  &  &  &  &  &  &  &  &  &  &  &  &  &  &  &  &  &  &  &  &  &  &  &  &  &  & \\
\hhline{~~------------------------------------------------}
 & fa & \cellcolor{red!10}{16} &  &  &  &  &  &  &  &  & \cellcolor{red!10}{85} &  &  &  &  &  &  &  &  &  &  &  &  &  &  &  &  &  &  &  &  &  &  &  &  &  &  &  &  &  &  &  &  &  &  &  &  &  & \\
\hhline{~~------------------------------------------------}
 & fi &  &  &  &  &  &  &  &  &  & \cellcolor{red!10}{79} &  &  &  &  &  &  &  &  &  &  &  &  &  &  &  &  &  &  &  &  &  &  &  &  &  &  &  &  &  &  &  &  &  &  &  &  &  & \\
\hhline{~~------------------------------------------------}
 & fr &  &  &  &  &  &  &  & \cellcolor{red!25}{683} &  & \cellcolor{red!40}{2868} & \cellcolor{red!10}{78} &  &  &  &  &  &  &  &  &  &  &  &  &  & \cellcolor{red!25}{403} & \cellcolor{red!10}{59} &  &  & \cellcolor{red!25}{344} &  &  &  &  &  & \cellcolor{red!10}{46} &  &  &  &  &  &  &  &  &  &  &  &  & \cellcolor{red!10}{61}\\
\hhline{~~------------------------------------------------}
 & ga &  &  &  &  &  &  &  &  &  & \cellcolor{red!10}{17} &  &  &  &  &  &  &  &  &  &  &  &  &  &  &  &  &  &  &  &  &  &  &  &  &  &  &  &  &  &  &  &  &  &  &  &  &  & \\
\hhline{~~------------------------------------------------}
 & gl &  &  &  &  &  &  &  &  &  & \cellcolor{red!10}{70} &  &  &  &  &  &  &  &  &  &  &  &  &  &  &  &  &  &  &  &  &  &  &  &  &  &  &  &  &  &  &  &  &  &  &  &  &  & \\
\hhline{~~------------------------------------------------}
 & he &  &  &  &  &  &  &  &  &  & \cellcolor{red!10}{59} &  &  &  &  &  &  &  &  &  &  &  &  &  &  &  &  &  &  &  &  &  &  &  &  &  &  &  &  & \cellcolor{red!10}{51} &  &  &  &  &  &  &  &  & \\
\hhline{~~------------------------------------------------}
 & hi &  & \cellcolor{red!10}{8} &  &  &  &  &  &  &  & \cellcolor{red!25}{367} &  &  &  &  &  &  &  &  &  &  &  &  &  &  &  &  &  &  &  &  &  &  &  &  &  &  &  &  &  &  &  &  &  &  &  &  &  & \\
\hhline{~~------------------------------------------------}
 & hr &  &  &  &  &  &  &  &  &  & \cellcolor{red!10}{81} &  &  &  &  &  &  &  &  &  &  &  &  &  &  &  &  &  & \cellcolor{red!10}{29} &  &  &  &  &  &  &  &  &  &  &  &  &  &  &  &  &  &  &  & \\
\hhline{~~------------------------------------------------}
 & hu &  &  &  &  &  &  &  &  &  & \cellcolor{red!10}{53} &  &  &  &  &  &  &  &  &  &  &  &  &  &  &  &  &  &  &  &  &  &  &  &  &  &  &  &  &  &  &  &  &  &  &  &  &  & \\
\hhline{~~------------------------------------------------}
 & hy &  &  &  &  &  &  &  &  &  & \cellcolor{red!10}{48} &  &  &  &  &  &  &  &  &  &  &  &  &  &  &  &  &  &  &  &  &  &  &  &  &  &  &  &  &  &  &  &  &  &  &  & \cellcolor{red!10}{13} &  & \\
\hhline{~~------------------------------------------------}
 & id &  &  &  &  &  &  &  &  &  & \cellcolor{red!10}{63} &  &  &  &  &  &  &  &  &  &  &  &  &  &  &  &  &  &  &  &  &  &  &  &  &  &  &  &  &  &  &  &  &  &  &  &  &  & \\
\hhline{~~------------------------------------------------}
 & it &  &  &  &  &  &  &  &  &  & \cellcolor{red!25}{801} &  &  &  &  &  &  &  &  &  &  &  &  &  &  &  &  &  &  &  &  &  &  &  &  &  &  &  &  &  &  &  &  &  &  &  &  &  & \\
\hhline{~~------------------------------------------------}
 & ja &  &  &  &  &  &  &  & \cellcolor{red!10}{60} &  & \cellcolor{red!25}{912} & \cellcolor{red!10}{67} &  &  &  & \cellcolor{red!25}{122} &  &  &  &  &  &  &  &  &  &  & \cellcolor{red!25}{163} &  &  &  &  &  &  &  &  &  &  &  &  &  &  &  &  &  &  &  &  &  & \cellcolor{red!10}{74}\\
\hhline{~~------------------------------------------------}
 & ko &  &  &  &  &  &  &  & \cellcolor{red!10}{70} &  & \cellcolor{red!40}{1004} & \cellcolor{red!10}{72} &  &  &  & \cellcolor{red!25}{110} &  &  &  &  &  &  &  &  &  & \cellcolor{red!25}{358} &  &  &  &  &  &  &  &  &  &  &  &  &  &  &  &  &  &  &  &  &  &  & \cellcolor{red!10}{73}\\
\hhline{~~------------------------------------------------}
 & lt &  &  &  & \cellcolor{red!10}{28} &  &  &  &  &  & \cellcolor{red!10}{68} &  &  &  &  &  &  &  &  &  &  &  &  &  &  &  &  &  &  &  &  &  &  &  &  &  &  &  &  &  &  &  &  &  &  &  &  &  & \\
\hhline{~~------------------------------------------------}
 & lv &  &  &  &  &  &  &  &  &  & \cellcolor{red!10}{61} &  &  &  &  &  &  &  &  &  & \cellcolor{red!10}{24} &  &  &  &  &  &  &  &  &  &  &  &  &  &  &  &  &  &  &  &  &  &  &  &  &  &  &  & \\
\hhline{~~------------------------------------------------}
 & mr &  &  &  &  &  &  &  &  &  & \cellcolor{red!10}{63} &  &  &  &  &  &  &  &  &  &  &  &  &  &  &  &  &  &  &  &  &  &  &  &  &  &  &  &  &  &  &  &  &  &  &  &  &  & \\
\hhline{~~------------------------------------------------}
 & nl &  &  &  &  &  &  &  &  &  & \cellcolor{red!10}{73} &  &  &  &  &  &  &  &  &  &  &  &  &  &  &  &  &  &  &  &  &  &  &  &  &  &  &  &  &  &  &  &  &  &  &  &  &  & \\
\hhline{~~------------------------------------------------}
 & no &  &  &  &  &  &  &  &  &  & \cellcolor{red!10}{53} &  &  &  &  &  &  &  &  &  &  &  &  &  &  &  &  &  &  &  &  &  &  &  &  &  &  &  &  &  &  &  &  &  &  &  &  &  & \\
\hhline{~~------------------------------------------------}
 & pl &  &  &  &  &  &  &  &  &  & \cellcolor{red!10}{65} &  &  &  &  &  &  &  &  &  &  &  &  &  &  &  &  &  &  & \cellcolor{red!25}{111} &  &  &  &  &  &  & \cellcolor{red!10}{58} &  &  &  &  &  &  &  &  &  &  &  & \\
\hhline{~~------------------------------------------------}
 & pt &  &  &  &  &  &  &  &  &  & \cellcolor{red!10}{89} &  &  &  &  &  &  &  &  &  &  &  &  &  &  &  &  &  &  &  &  &  &  &  &  &  &  &  & \cellcolor{red!10}{40} &  &  &  &  &  &  &  &  &  & \\
\hhline{~~------------------------------------------------}
 & ro &  &  &  &  &  &  &  &  &  & \cellcolor{red!10}{91} &  &  &  &  &  &  &  &  &  &  &  &  &  &  &  &  &  &  &  &  &  &  &  &  &  &  &  &  &  &  &  &  &  &  &  &  &  & \\
\hhline{~~------------------------------------------------}
 & ru &  &  &  &  &  &  &  & \cellcolor{red!25}{106} &  & \cellcolor{red!25}{472} & \cellcolor{red!10}{87} &  &  &  & \cellcolor{red!10}{42} &  &  &  &  &  &  &  &  &  &  &  &  &  &  &  &  &  &  &  &  &  &  &  &  &  &  &  &  &  &  &  &  & \\
\hhline{~~------------------------------------------------}
 & sk &  &  &  &  &  &  &  &  &  & \cellcolor{red!10}{54} &  &  &  &  &  &  &  &  &  &  &  &  &  &  &  &  &  &  &  &  &  & \cellcolor{red!10}{17} &  &  &  &  &  &  &  &  &  &  &  &  &  &  &  & \\
\hhline{~~------------------------------------------------}
 & sl &  &  &  &  &  &  &  &  &  & \cellcolor{red!10}{69} &  &  &  &  &  &  &  &  &  &  &  &  &  &  &  &  &  &  &  &  &  &  &  &  &  &  &  &  &  &  &  &  &  &  &  &  &  & \\
\hhline{~~------------------------------------------------}
 & sr &  &  &  &  &  &  &  &  &  & \cellcolor{red!10}{64} &  &  &  &  &  &  &  &  &  &  &  &  &  &  &  &  &  &  &  &  &  &  & \cellcolor{red!10}{54} &  &  &  &  &  &  &  &  &  &  &  &  &  &  & \\
\hhline{~~------------------------------------------------}
 & sv &  &  &  &  &  &  &  &  &  & \cellcolor{red!10}{79} &  &  &  &  &  &  &  & \cellcolor{red!10}{28} &  &  &  &  &  &  &  &  &  &  &  &  &  &  &  &  &  &  &  &  &  &  &  &  &  &  &  &  &  & \\
\hhline{~~------------------------------------------------}
 & sw &  &  &  &  &  &  &  &  &  & \cellcolor{red!25}{327} &  &  &  &  &  &  &  &  &  &  &  &  &  &  &  &  &  &  &  &  &  &  &  &  &  &  &  &  &  &  &  &  &  &  &  &  &  & \\
\hhline{~~------------------------------------------------}
 & ta &  &  &  &  &  &  &  &  &  & \cellcolor{red!10}{39} &  &  &  &  &  &  &  &  &  &  &  &  &  &  &  &  &  &  &  &  &  &  &  &  &  &  &  &  &  &  &  &  &  &  &  &  &  & \\
\hhline{~~------------------------------------------------}
 & th &  &  &  &  &  &  &  &  &  & \cellcolor{red!25}{299} &  &  &  &  &  &  &  &  &  &  &  &  &  &  &  &  &  &  &  &  &  &  &  &  &  &  &  &  &  &  &  &  &  &  &  &  &  & \\
\hhline{~~------------------------------------------------}
 & tr &  &  &  &  &  &  &  &  &  & \cellcolor{red!25}{386} &  &  &  &  &  &  &  &  &  &  &  &  &  &  &  &  &  &  &  &  &  &  &  &  &  &  &  &  &  &  &  &  &  &  &  &  &  & \\
\hhline{~~------------------------------------------------}
 & uk &  &  &  &  &  &  &  &  &  & \cellcolor{red!10}{77} &  &  &  &  &  &  &  &  &  &  &  &  &  &  &  &  &  &  &  &  &  &  &  &  &  &  &  &  &  &  &  &  &  &  &  &  &  & \\
\hhline{~~------------------------------------------------}
 & ur &  &  &  &  &  &  &  &  &  & \cellcolor{red!25}{372} &  &  &  &  &  &  &  &  &  &  &  &  &  &  &  &  &  &  &  &  &  &  &  &  &  &  &  &  &  &  &  &  &  &  &  &  &  & \\
\hhline{~~------------------------------------------------}
 & vi &  &  &  &  &  &  &  &  &  & \cellcolor{red!25}{391} &  &  &  &  &  &  &  &  &  &  &  & \cellcolor{red!10}{3} &  &  &  &  &  &  &  &  &  &  &  &  &  &  &  &  &  &  &  &  &  &  &  &  &  & \\
\hhline{~~------------------------------------------------}
 & wo &  &  &  &  &  &  &  &  &  & \cellcolor{red!10}{11} &  &  &  &  &  &  &  &  &  &  &  &  &  &  &  &  &  &  &  &  &  &  &  &  &  &  &  &  &  &  &  &  &  &  &  &  &  & \\
\hhline{~~------------------------------------------------}
 & zh &  &  &  &  &  &  &  & \cellcolor{red!25}{150} &  & \cellcolor{red!40}{1209} & \cellcolor{red!10}{59} &  &  &  & \cellcolor{red!25}{113} &  &  &  &  &  &  &  &  &  & \cellcolor{red!25}{128} & \cellcolor{red!10}{80} &  &  &  &  &  &  &  &  &  &  &  &  &  &  &  &  &  &  &  &  &  & \\
\hhline{~~------------------------------------------------}
\end{tabular}
\caption{Number of examples per language pair. Rows: source language; Columns: target language}
\label{lang_pair_matrix}
\end{sidewaystable*}

% 1-99: no colour
% 100-999: pale
% 1,000-9,999: mid
% 10,000+: dark
Table~\ref{lang_pair_matrix} contains the total number of examples per language pair in the challenge set. As can be seen in the table, the distribution of examples is variable across language pairs. The dominant language pairs are: en-de, de-en, and fr-en.

\subsection{Distribution of Language Pairs Across Phenomena}
\label{app:language_pair_phenomena}
\begin{sidewaystable*}[]
\tiny

\begin{tabular}{p{0.24\linewidth} | p{0.23\linewidth} | p{0.19\linewidth} | p{0.23\linewidth}}
\hline
phenomena &
  language pairs &
  phenomena &
  language pair \\ \hline
\begin{tabular}[c]{@{}l@{}}ambiguous-translation-wrong-\\ discourse-connective-since-causal\\
ambiguous-translation-wrong-\\ discourse-connective-since-temporal\\hallucination-unit-conversion-unit-matches-ref\end{tabular} &
  fr-en, de-en &
  hallucination-real-data-vs-ref-word &
  en-de, de-en, fr-de \\ \hline
ambiguous-translation-wrong-discourse-connective-while-contrast &
  fr-en &
  hallucination-real-data-vs-synonym &
  en-mr, de-en, en-de, fr-de \\ \hline
ambiguous-translation-wrong-discourse-connective-while-temporal &
  fr-en &
  untranslated-vs-ref-word &
  en-de, de-en, fr-de \\ \hline
ambiguous-translation-wrong-gender-female-anti &
  fr-en, de-en, it-en &
  untranslated-vs-synonym &
  en-de, de-en, fr-de \\ \hline
ambiguous-translation-wrong-gender-male-anti &
  fr-en, de-en, it-en &
  modal\_verb:deletion &
  de-en \\ \hline
ambiguous-translation-wrong-gender-male-pro &
  fr-en, de-en, it-en &
  modal\_verb:substitution &
  de-en \\ \hline
ambiguous-translation-wrong-sense-frequent &
  en-de, en-ru &
  nonsense &
  ko-en, ko-ja, en-ko, fr-ja, de-en \\ \hline
ambiguous-translation-wrong-sense-infrequent &
  en-de, en-ru &
  ordering-mismatch &
  en-de, de-en, fr-de \\ \hline
anaphoric\_group\_it-they:deletion &
  en-de &
  overly-literal-vs-correct-idiom &
  en-de, de-en \\ \hline
anaphoric\_group\_it-they:substitution &
  en-de &
  overly-literal-vs-explanation &
  en-de, de-en \\ \hline
anaphoric\_intra\_non-subject\_it:deletion &
  en-de &
  overly-literal-vs-ref-word &
  en-de, de-en, fr-de \\ \hline
anaphoric\_intra\_non-subject\_it:substitution &
  en-de &
  overly-literal-vs-synonym &
  en-mr, de-en, en-de, fr-de \\ \hline
anaphoric\_intra\_subject\_it:deletion &
  en-de &
  pleonastic\_it:deletion &
  en-de \\ \hline
anaphoric\_intra\_subject\_it:substitution &
  en-de &
  pleonastic\_it:substitution\_pro\_trans\_different\_to\_ref &
  en-de \\ \hline
anaphoric\_intra\_they:deletion &
  en-de &
  punctuation:deletion\_all &
  en-de \\ \hline
anaphoric\_intra\_they:substitution &
  en-de &
  punctuation:deletion\_commas &
  en-de \\ \hline
anaphoric\_singular\_they:deletion &
  en-de &
  punctuation:deletion\_quotes &
  en-de \\ \hline
anaphoric\_singular\_they:substitution &
  en-de &
  \begin{tabular}[c]{@{}l@{}}punctuation:statement-to-question\\ do-not-translate\end{tabular}
   &
  en-de \\ \hline
antonym-replacement &
  fr-en, ko-en, ja-en, es-en, zh-en, de-en &
  real-world-knowledge-entailment &
  en-de, de-en \\ \hline
similar-language-high &
  en-hi, en-cs, en-es &
  real-world-knowledge-hypernym-vs-distractor &
  en-de, de-en \\ \hline
similar-language-low &
  fr-mr, en-pl, en-ca &
  real-world-knowledge-hypernym-vs-hyponym &
  en-de, de-en \\ \hline
\begin{tabular}[c]{@{}l@{}}coreference-based-\\ on-commonsense\end{tabular} &
  en-de, en-ru, en-fr &
  real-world-knowledge-synonym-vs-antonym &
  en-de, de-en \\ \hline
\begin{tabular}[c]{@{}l@{}}hallucination-named-entity-level-1\\ hallucination-named-entity-level-2\\ hallucination-named-entity-level-3\\ hallucination-number-level-1\\ hallucination-number-level-2\\ hallucination-number-level-3\end{tabular} &
  en-de, ja-de, en-ko, de-zh, ja-en, es-de, fr-en, es-ko, ko-ja, es-ja, de-ja, zh-es, fr-zh, fr-ja, es-en, fr-ko, zh-en, ko-de, ko-es, de-ko, ko-en, fr-es, ja-es, ja-ko, zh-fr, en-es, de-en, ja-fr, ko-zh, en-fr, de-fr, ko-fr, es-fr, zh-ko, fr-de, ja-zh, de-es, es-zh, en-ja, zh-de, en-zh, zh-ja &
  \begin{tabular}[c]{@{}l@{}}undertranslation\\ overtranslation\end{tabular} &
  fr-en, ko-en, ja-en, es-en, zh-en, de-en \\ \hline
lexical-overlap &
  fr-en, en-fr, de-fr, ko-en, es-ja, ja-en, ko-fr, es-fr, ko-ja, de-ja, zh-en, ja-fr, zh-fr, en-ja, es-en, fr-ja, de-en, zh-ja &
  \begin{tabular}[c]{@{}l@{}}xnli-addition-contradiction\\ xnli-addition-neutral\\ xnli-omission-contradiction\\ xnli-omission-neutral\end{tabular} &
  fr-en, vi-en, sw-en, tr-en, zh-en, ru-en, bg-en, el-en, th-en, es-en, hi-en, de-en, ar-en, ur-en \\ \hline
\begin{tabular}[c]{@{}l@{}}hallucination-unit-conversion-amount-matches-ref\\ hallucination-unit-conversion-unit-matches-ref\end{tabular} &
  et-en, wo-en, da-en, no-en, uk-en, ta-en, fi-en, pl-en, ja-en, hy-en, ur-en, hr-en, fr-en, lt-en, tr-en, he-en, bg-en, ro-en, sv-en, ru-en, es-en, nl-en, zh-en, hu-en, be-en, lv-en, ko-en, ga-en, sk-en, af-en, sl-en, sr-en, ca-en, de-en, mr-en, id-en, vi-en, gl-en, pt-en, fa-en, hi-en, el-en, ar-en, it-en, cs-en &
  hallucination-date-time &
  en-de, et-en, ca-es, en-et, hr-lv, da-en, no-en, uk-en, fi-en, en-da, ta-en, pl-en, ja-en, en-hr, hy-en, ur-en, fr-en, hr-en, lt-en, sr-pt, en-sv, tr-en, en-no, en-sl, he-en, pl-sk, ru-en, ro-en, sv-en, en-lt, es-en, en-nl, nl-en, bg-en, he-sv, zh-en, hu-en, be-en, lv-hr, lv-en, bg-lt, en-ro, sk-pl, ko-en, ga-en, sk-en, af-en, sl-en, en-hu, sr-en, en-es, ca-en, en-sk, de-en, mr-en, id-en, vi-en, gl-en, en-fr, de-fr, pt-en, fr-de, en-pt, fa-en, hi-en, el-en, ar-en, it-en, en-pl, cs-en \\ \hline
\begin{tabular}[c]{@{}l@{}}commonsense-only-ref-ambiguous\\ commonsense-src-and-ref-ambiguous\end{tabular} &
  en-de, fr-en, ru-fr, en-fr, de-fr, ru-de, fr-de, ru-en, en-ru, fr-ru, de-ru, de-en &
  copy-source &
  ar-fr, ru-es, ur-en, fr-en, tr-en, zh-de, bg-en, ru-en, es-en, zh-en, sw-en, ja-ko, th-en, de-en, pl-mr, vi-en, hi-en, el-en, ar-en \\ \hline
\begin{tabular}[c]{@{}l@{}}addition\\ omission\end{tabular} &
  en-ca, en-el, en-et, en-ta, pl-en, hr-en, he-en, pl-sk, en-ar, ru-en, en-fi, zh-en, hu-en, be-en, lv-hr, en-he, ko-en, en-fa, sl-en, ca-en, en-gl, en-tr, en-sk, de-en, en-sr, fa-af, fa-en, ar-en, cs-en, en-de, en-hy, ar-hi, no-en, uk-en, fi-en, en-be, sr-pt, en-ru, sv-en, nl-en, sk-pl, en-hi, en-hu, mr-en, hi-ar, id-en, gl-en, en-fr, en-lv, fr-de, ca-es, en-uk, &
  \begin{tabular}[c]{@{}l@{}}addition\\ omission\end{tabular} &
  en-ur, en-hr, ur-en, en-no, en-sl, ro-en, en-vi, en-lt, es-en, en-nl, he-sv, en-it, en-ro, af-fa, en-id, lt-bg, en-af, af-en, es-ca, vi-en, sv-he, de-fr, pt-en, en-pl, et-en, hr-lv, wo-en, da-en, en-ko, en-da, ja-en, hy-en, pt-sr, hy-vi, fr-en, en-cs, lt-en, en-sv, tr-en, bg-en, lv-en, bg-lt, sr-en, en-es, en-bg, en-pt, hi-en, el-en, it-en \\ \hline
\end{tabular}
\caption{Collection of list of languages per phenomena}
\label{tab:lang-phenomena}

\end{sidewaystable*}

Table~\ref{tab:lang-phenomena} contains the list of language pairs per phenomena in the challenge set. As can be seen in the table, the distribution of language pairs is variable across phenomena. Addition and omission have the highest variety of language pairs. en-de is the most frequent language pair across all phenomena.

\end{document}